\documentclass[3p,twocolumn,review,times]{elsarticle}

\usepackage{amssymb}

\usepackage{lipsum} 
\usepackage{adjustbox}
\usepackage{multirow}
\usepackage{comment}
\usepackage{xurl}
\usepackage{parskip}
\usepackage{float}

\usepackage[table]{xcolor}
\usepackage{tikz}
\usepackage{collcell}

\usepackage{xcolor}

\newcommand*\rot{\rotatebox{90}}

\def\ccp#1{%
    \pgfmathsetmacro\calc{(#1)*100/(100)}%
    \edef\clrmacro{\noexpand\cellcolor{blue!\calc}}%
    \clrmacro%
    \ifdim \calc pt>50pt\color{white}\fi{#1}%
}

\def\ccn#1{%
    \pgfmathsetmacro\calc{(#1)*100/(100)}%
    \edef\clrmacro{\noexpand\cellcolor{red!\calc}}%
    \clrmacro%
    \ifdim \calc pt>50pt\color{white}\fi{#1}%
}

\journal{Online Social Networks and Media}

\begin{document}

\begin{frontmatter}

\title{Negativity Spreads Faster: A Large-Scale Multilingual Twitter Analysis \\ on the Role of Sentiment in Political Communication}


\author{Dimosthenis Antypas, Alun Preece, Jose Camacho-Collados}
\ead{\{antypasd, preecead, camachocolladosj\}@cardiff.ac.uk}

\address{Cardiff University, School of Computer Science and Informatics, Cardiff NLP \& Crime and Security Research Institute, Cardiff, CF24 3AA, Wales, United Kingdom}

\begin{abstract}

Social media has become extremely influential when it comes to policy making in modern societies, especially in the western world, where platforms such as Twitter allow users to follow politicians, thus making citizens more involved in political discussion. In the same vein, politicians use Twitter to express their opinions, debate among others on current topics and promote their political agendas aiming to influence voter behaviour. In this paper, we attempt to analyse tweets of politicians from three European countries and explore the virality of their tweets.
Previous studies have shown that tweets conveying negative sentiment are likely to be retweeted more frequently. By utilising state-of-the-art pre-trained language models, we performed sentiment analysis on hundreds of thousands of tweets collected from members of parliament in Greece, Spain and the United Kingdom, including devolved administrations. We achieved this by systematically exploring and analysing the differences between influential and less popular tweets. 
Our analysis indicates that politicians’ negatively charged tweets spread more widely, especially in more recent times, and highlights interesting differences between political parties as well as between politicians and the general population.

\end{abstract}

\begin{keyword}
Politics\sep Twitter\sep NLP


\end{keyword}

\end{frontmatter}


\section{Introduction}\label{}
In recent years social media have come to resemble a `battleground' between politicians who constantly aim to reach out to people and win their votes. This behaviour is not surprising as the influx of people onto social media aiming to get updated on the latest news keeps growing, especially in the west, with 48\% of Europeans using social media on a regular basis \cite{cattaneo_2020}. More specifically, Twitter seems to have become established as the main online platform where politicians attempt to engage with the public in regards to social and political commenting \cite{stier2018election}, to such an extent that political accounts are often more active than non-political ones \cite{grant2010digital}. Due to their inherent interest and thanks to the openness of Twitter's API, political tweets have been extensively studied. Using tweets extracted from politicians and their followers, researchers have attempted to answer questions such as: is it possible to predict elections results \cite{tumasjan2010predicting}? or what makes a politician popular on social media \cite{vaccari2013drives}? 

In general, discovering what makes a message viral has been a popular research topic. Sentiment analysis is one of the tools that have been extensively used to answer this question. Findings in related literature suggest that negatively charged tweets have a bigger network penetration than average \cite{tsugawa2017relation,naveed2011bad,jimenez2021sentiments}. However, even though there have been studies of sentiment in tweets revolving around politics/elections \cite{chung2011can,parmelee2011politics,llewellyn2016brexit}, to our knowledge there has not been a large-scale multilingual analysis of the relation between sentiment and the propagation of politicians' tweets.

In this paper, we focus on politicians' tweets to understand the relation between their sentiment and virality. By performing a more fine-grained analysis, a distinction is made between politicians from different political parties as we attempt to identify differences in their tweeting activities. At the same time, we investigate whether politicians' behaviour regarding their tweet sentiment is independent of their home country and language, and also whether the behaviour is consistent or evolves over time.

Overall, we bring together and assess the validity of these and other research questions (Section \ref{sec:rqs}) with regards to politicians and Twitter by: (1) collecting a large-scale dataset of recent tweets by members of parliament (MPs) in three different countries (Greece, Spain and United Kingdom) and multiple languages (Section \ref{sec:datacollect}); (2) establishing a robust evaluation making use of state-of-the-art multilingual sentiment analysis models powered by the recent successes of transformer-based language models (Section \ref{sec:models_evaluation}); and (3) performing a multi-faceted analysis to investigate relations between sentiment and virality of tweets (Section \ref{sec:analysis}).

In addition to the specific sections where our main research questions are addressed, we included relevant control experiments for robustness across different aspects in Section \ref{sec:control}. These control experiments help put in context the prior research questions and provide extra insights about those. Finally, all our findings linked to these research questions are summarised in Section \ref{sec:summary}.

\section{Research Questions}
\label{sec:rqs}

The aims of this paper are summarised by the following three research questions.

\paragraph{RQ1. What is the best sentiment analysis classifier when dealing with political content in social media?} This question is mainly answered in Section \ref{sec:models_evaluation}. To answer this question, we first constructed an ad hoc sentiment analysis dataset with Twitter messages of members of parliament in Greece, Spain and United Kingdom. Then, we compared a wide range of settings and sentiment analysis models, including language models. 

\paragraph{RQ2. Is there a correlation between virality and sentiment in political communication in social media?} This question is mainly answered in Section \ref{sec:sentvir}. To answer this question, we relied on state-of-the-art sentiment analysis models (as tested in RQ1) and analysed a large corpus of almost one million tweets comprising all Twitter posts by members of parliament from Greece, Spain and United Kingdom. Then, we ran various statistical tests to measure and understand the possible correlation between virality and sentiment.

\paragraph{RQ3. Is there a difference in the sentiment of social media posts between (a) government and opposition, and (b) devolved parliaments?} These questions are mainly answered in Section \ref{sec:opposition} (a) and Section \ref{sec:devolved} (b). To answer these questions, in addition to analyse the data by all MPs in the main governments, we extracted and analysed data split by political party and extended to members of devolved parliaments in Spain (Catalonia and Basque Country) and United Kingdom (Northern Ireland, Scotland and Wales).

\section{Related Work}
The popularity of social media platforms such as Facebook and Twitter is transforming the way citizens and politicians communicate with one another. Political candidates and voters use Twitter to discuss social and political issues, sharing information and encouraging political participation \cite{kushin2013did}. Politicians in particular, especially in recent years, have eagerly embraced social media tools to self-promote and communicate with their electorate, seeing in these tools the potential for changing public opinion especially during election campaigns \cite{hong2011does}. Given the rapid growth of politicians' engagement through Twitter, there is plenty of research on how the platform is used for political communication.

Many studies focus on the classification of tweets referring to politicians by sentiment --- positive, negative, or neutral --- to investigate popularity and voting intention, and whether there is a correlation between post sentiments and political results \cite{taddy2013measuring,vilares2015megaphone,kermanidis2013political}. Moreover,  sentiment is considered to affect message diffusion in social media. Research suggests that the virality of a message --- the probability of it being retweeted --- is affected by its polarity, as emotionally charged messages tend to be re-posted more rapidly and frequently compared to neutral ones \cite{stieglitz2013emotions}. Negative messages in particular are likely to be more retweeted than positive and neutral ones \cite{tsugawa2015negative}. However, other studies show that the relationship between sentiment and virality in Twitter is more complex and is related to subject area \cite{hansen2011good}. Literature suggests that sentiment occurring in politically relevant tweets has an effect on their propagation \cite{stieglitz2011role, stieglitz2012political}.

When considering techniques used to extract sentiment from political text in social media it is common to utilise dictionary based approaches \cite{karami2019political, heidenreich2020political, rathje2021out}  or in cases where the platform offers the functionality to react in one's post (e.g., Facebook's like/angry/happy reactions) then an aggregation of such reactions is used to determine the sentiment of the post \cite{sandoval2018facebook,sandoval2020sentiment}. These approaches have the benefit of not requiring to train machine learning models for the sentiment analysis task which itself can be a time consuming process while also requiring previously annotated data. Other researchers chose to train their own ML models instead and often utilise neural network architectures such as LSTM and GRU networks \cite{svetlov2019sentiment, dorle2018political}. Despite the variety of methods utilised, there seems to be a lack of usage of state-of-the-art NLP methods like language models such as BERT \cite{devlin_bert} and RoBERTa \cite{liu2019roberta}. In Section \ref{sec:models_evaluation}, we show how this makes an important difference in practice given the important improvement attained by language models in NLP in recent years.

Finally, politics and social media is a popular research area in multiple countries across the globe, particularly where platforms such as Twitter and Instagram have a strong presence. For example, studies focusing in the US political scene have tried to identify popularity measures \cite{peng2021makes}, trending topics \cite{karami2019political}, what drives public engagement \cite{rathje2021out} and relations between sentiment and politicians popularity \cite{joyce2017sentiment}. Similar research can be found in other countries such as UK  \cite{mee2021sentiment}, but also in non-English speaking countries such as Russia \cite{svetlov2019sentiment}, Mexico \cite{sandoval2020sentiment},  Italy \cite{ceron2016campaigning}, Germany \cite{stieglitz2012political}, Austria \cite{kuvsen2018politics}  as well as in cross-European settings  \cite{heidenreich2020political}.

\section{Data Collection}
\label{sec:datacollect}
For the purposes of this study, tweets were collected from the MPs (members of parliament) of three 
sovereign European countries, namely, the United Kingdom (UK),  Greece and Spain, including several of their devolved parliaments\footnote{We decided to include these devolved parliaments due to their size, idiosyncrasy and nationalist identity.}:
Northern Ireland, Scotland and Wales in the UK and the Basque Country and Catalonia in Spain for 2021.

\paragraph{Countries Selection} The above countries were chosen because even though they share similar characteristics (e.g., they are all European) they have distinct attributes that makes a comparison between them more interesting. These distinctions are both linguistic, where English, Spanish and Greek are not closely related languages, and also socioeconomic.

For example, the UK has a population of 67.33 million and a GDP per capita of 47,334\$ and is the largest and richest country of the group, followed by Spain with a population of 47.33 million and GDP per capita of 30,115\$, and finally Greece which is the least populated (10.67 million) and with a smaller economy (20,276\$ GDP per capita) \footnote{Demographic data extracted from: https://data.worldbank.org/}. At the same time, the three countries seem to hold different views to social subjects such as religion \cite{pew2018religious} and LGBTI inclusion \cite{flores2021social}. Socioeconomic differences as the above, have been shown to affect public perception on national issues (e.g., sense of solidarity \cite{vasilopoulou2020poor}) but also the way politicians engage with the public \cite{brown2017politics}.

As a final note, due to the nature of the analysis, it is beneficial that our research team includes researchers that have a good understanding of the languages as well as awareness of the political environments of the countries studied.

\paragraph{Time Period Selection} Aiming to minimise any potential biases that could occur during election periods, January to December 2021 was chosen as the time period for our main analysis. The last national elections for the three main parliaments took place in 2019 and thus we can assume that any campaigning rhetoric will be absent from the tweets. Finally, this time period is particularly interesting as all three countries had to deal with the common issue of the COVID-19 pandemic.

\paragraph{Control Data} In order to establish relevant comparison points with respect to the specific time period examined and the general population, we collected additional Twitter corpora. These included (1) tweets from MPs from the UK's national (London) parliament for 2014 \& 2016, (2) tweets from random users from Greece, UK and Spain and (3) tweets from verified users residing in the aforementioned countries (see Section \ref{sec:controldat} for more details).

\paragraph{Collection Method} In total 2,933,143 tweets were scraped from 157,333 users of which 2,213 represented members of the parliaments for the aforementioned countries. The total number of tweets acquired from politicians were 1,588,970. The collection of the tweets was achieved using Twitter's API while utilising Python's Tweepy \cite{roesslein2009tweepy} and Twarc \cite{twarc} libraries. Retweets were ignored; only original tweets were considered, as the main metric to measure the popularity of a tweet (the retweet count) is not available on retweeted messages, which is crucial for our sentiment exploration. Finally, tweets that do not contain  meaningful text, e.g., text contains only URLs, were also discarded.

\subsection{Main 2021 MP Twitter Corpus}
\label{maindataset}

\begin{figure*}[th]
\centering
\begin{adjustbox}{width=\textwidth,center}
\includegraphics{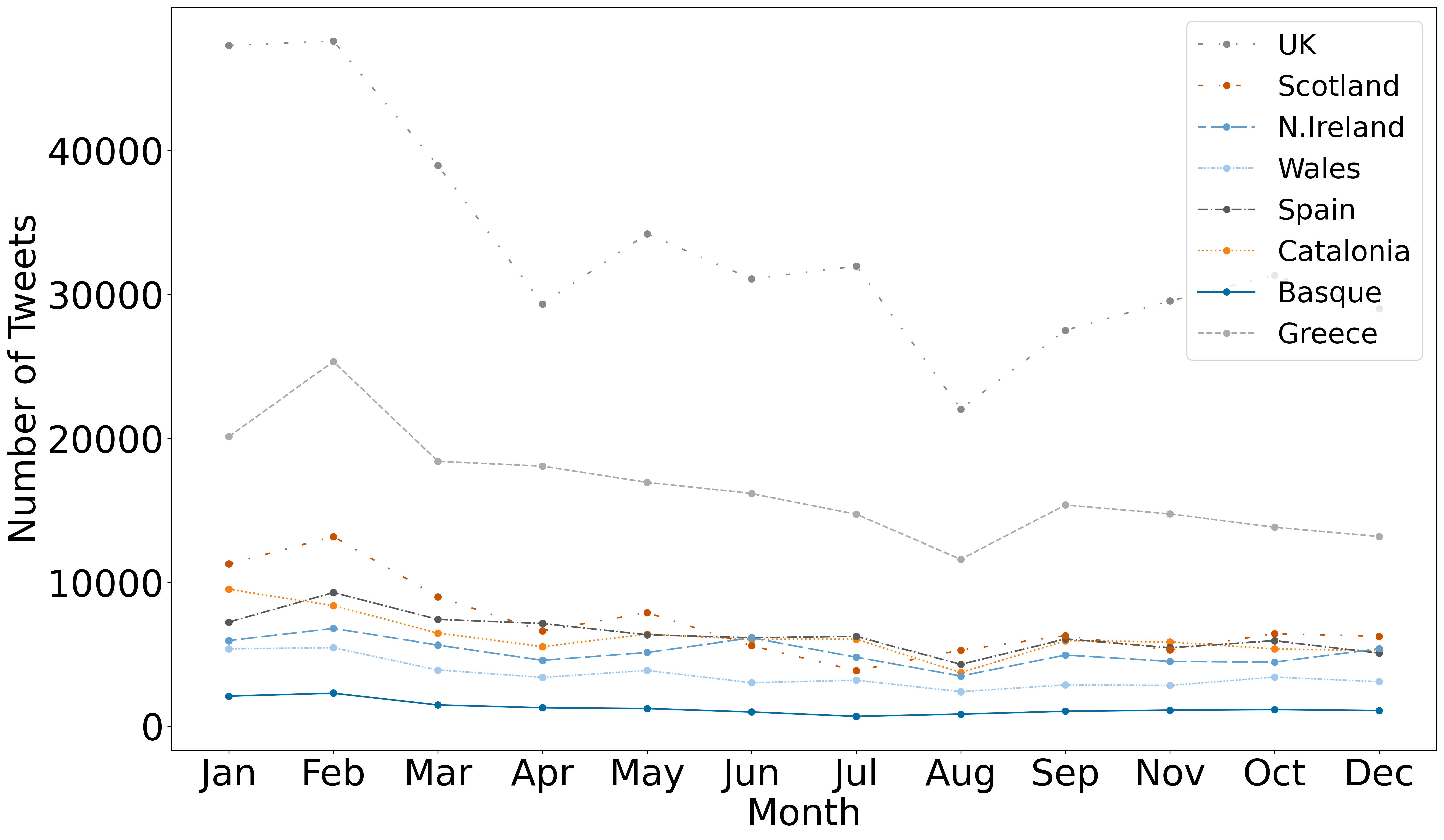}
\end{adjustbox}
\caption{Number of MP tweets collected through 2021.}
\label{fig:tweets_collected}
\end{figure*}

To compile our 2021 dataset, we extracted tweets from the members of parliament of the three sovereign countries analysed, i.e., Greece, Spain and UK and their devolved parliaments. Figure \ref{fig:tweets_collected} displays the total number of tweets collected for each of the months under study for all considered parliaments\footnote{Table \ref{tab:tweets_collected} of the Appendix provides more detailed results}.

\subsubsection{Parliaments of Greece, Spain and UK}\label{sec:data_2021} Our main analysis focuses on tweets from members of the UK's, Spain's, and Greece's parliaments from January to December 2021, will be referred to as: \textit{2021 Main Dataset}. For this time period, we scraped a  continuous  collection  of  tweets from 1,040 members of parliaments (UK: 577, Spain: 279, Greece: 184). The Twitter accounts of the MPs were manually retrieved and verified using either their respective parliament official website or Google search. There are several cases where MPs do not own a Twitter account or they have a protected account which makes the retrieval of their tweets impossible. Consequently, our dataset is not necessarily proportional to the actual parliaments distributions, e.g., in the UK the governing Conservative party has a 56\% of the total seats whereas in our UK dataset the Conservative party represents 54\% of the total MPs. Finally, it is worth noting that Greek MPs tend to be the least involved with Twitter with only 61\% of them actually having an active account in contrast to the UK's and Spanish MPs with 96\% and 85\% active accounts respectively. 

We further analyse the data gathered and explore text features that according to previous research  \cite{pancer2016popularity,nanath2021leveraging} are related to the virality of tweets. 
Table \ref{tab:control_stats} displays text statistics extracted from tweets of the three main parliaments. Specifically, the percentage of tweets that include emojis, hashtags, mentions, and URLs as well as the percentage of upper case characters are shown. The overall majority of the tweets (68\%) contain at least one URL which is not surprising as sharing news and commenting on them is one of the main use case of Twitter. It is interesting to note that Greek MPs tend to use fewer emojis than their UK and Spain counterparts (44\% and 40\% less), while they include hashtags more often (44\%). The above statistics are going to be utilized in our regression analysis in Section \ref{sec:sentvir}.

\begin{table}[ht]
    \centering
    \begin{tabular}{lccccc}
    \textbf{Country} & \textbf{\#} & \textbf{@} & \textbf{emoji} & \textbf{url} & \textbf{up/low} \\ \hline
    UK               & 0.05        & 0.23       & 0.23           & 0.66         & 0.10            \\ \hline
    Spain            & 0.05        & 0.33       & 0.25           & 0.70         & 0.11            \\ \hline
    Greece           & 0.09        & 0.27       & 0.14           & 0.72         & 0.12            \\ \hline
    \textbf{Overall} & 0.05        & 0.27       & 0.23           & 0.68         & 0.11           
    \end{tabular}
    \caption{Percentage of tweets that include  hashtags (\#), mentions (@), emojis, and URLs in the UK, Greece, Spain 2021 main parliaments. The upper to low case ratio (up/low) average of tweets texts is also reported.}
    \label{tab:control_stats}
\end{table}

\subsubsection{Devolved Parliaments}\label{sec:devolved_2021}
Tweets from members of the devolved parliaments of the UK (N.Ireland, Scotland, Wales) and Spain (Basque Country, Catalonia) were also collected for the same time period.
These subsets, will be referred to as: \textit{2021 Devolved Dataset}, are used to identify potential differences in tweets between the main and the devolved parliaments regarding the prevailing sentiments. Again, a manual search and verification was applied for every MP in order to retrieve his/her Twitter handle. Specifically for Catalonia, care was taken when aggregating the MPs handles due to the elections that took place on February 14\textsuperscript{th}, 2021. Text statistics for tweets gathered from the devolved parliaments are presented in Table \ref{tab:control_stats_devolved} of the Appendix.

\subsection{Control Datasets}
\label{sec:controldat}

\subsubsection{UK parliament 2014 \& 2016}\label{sec:temporal_data}

Aiming to explore the sentiment trends we added a temporal element in our analysis by collecting tweets from UK's MPs for the years of 2014 and 2016, will be referred as the \textit{UK 2014 \& 2016} dataset. In this case, the MPs and their respective handles were collected by utilising the SPARQL \cite{SPARQL1113:online} endpoint  of Wikidata \cite{wikidata}. In total 202,954 and 429,538 tweets were extracted for 2014 and 2016 respectively. It should be mentioned that the number of Twitter handles scraped for each year was smaller than that of 2021; 156 and 474 handles for 2014 and 2016 respectively in comparison to 577 for the 2021 UK dataset. The increased popularity of Twitter over the years can potentially explain the larger number of handles for 2016 and 2021 over 2014. However, this might also be due to limitations of the Wikidata archive. 

\subsubsection{Random \& Verified Users 2021}\label{sec:random&verified} To compare politicians' tweets with those of the general population for 2021, two distinct sets of tweets of random and verified\footnote{According to Twitter, an account is deemed verified if it is authentic, notable and active (https://help.twitter.com/en/managing-your-account/about-twitter-verified-accounts).} users were also collected (each for any of the country's parliament studied --- Spain, Greece, the UK), will be referred  to as \textit{Random \& Verified 2021}. Verified users accounts usually belong to recognisable figures such as brands, organisations, and influential persons (e.g., athletes and artists) whose Twitter activity can be deemed to be closer to that of MPs. Each set of random users follows the same distribution of tweets of their respective country shown in Figure \ref{fig:tweets_collected}. The geolocalization of tweets was achieved using Twitter's API country filter (`place\_country'). An assumption was made that tweets belong to users that reside to the country they are posting from.
The verified users set was constructed by extracting tweets only from a list of known verified users and a combination of keywords\footnote{https://github.com/cardiffnlp/politics-and-virality-twitter/blob/main/data/keywords.csv} alongside with information on the location from user profiles metadata was applied to ensure the accounts resided in the countries studied.

\section{Sentiment Analysis: Evaluation and Model Selection}\label{sec:models_evaluation}

Over the years there have been multiple approaches of dealing with sentiment analysis in text data. Varying from the use of sentiment lexicons \cite{inui-yamamoto-2011-applying, taboada2011lexicon,banea2008bootstrapping} to utilising linear machine learning models \cite{ahmad2017sentiment} and, most recently, by applying transformer models like BERT \cite{devlin_bert}. One of the most challenging problems appears when we have to deal with multilingual data. Acquiring a model that is able to perform well in a multilingual setting is a difficult task that often requires large labelled corpora. This is especially true if low resource languages are taken into consideration \cite{barnes2018bilingual}. Some cross-lingual approaches, such as language models, deal with this issue by making use of the large amount of training data available in major languages, e.g., English, to essentially transfer sentiment information to low resource languages, e.g., Greek \cite{can2018multilingual}. It is also important to note that despite the architecture being used, an important factor to achieve accurate sentiment classification is the domain of the training and the target corpus \cite{peng2018cross}. 

For our purposes, we select a number of pre-trained language models, both monolingual and multilingual, which we attempt to further finetune and evaluate them using a manually labelled tweets dataset (Section \ref{sec:annotation}) aiming to find the best suitable classifier for each language (English, Spanish, Greek) studied.

\subsection{Sentiment Annotation}\label{sec:annotation}
Tweets from the \textit{2021 Main Dataset} (Section \ref{sec:data_2021}) for each language included in the study were sampled from their respective parliaments. This way three datasets were collected and annotated based on their sentiment for the English, Spanish and Greek languages (will be referred to as: \textit{Annotated Set}).

In sentiment annotation tasks, annotators are asked to either evaluate the overall polarity of the text on a scale, e.g., 1 to 5 \cite{al2017arasenti} or to distinct positive/neutral/negative classes \cite{patwa2020semeval}. For simplicity and to follow current state-of-the-art sentiment analysis models \cite{nguyen2020bertweet,barbieri2021xlm}, in our setting annotators were asked to indicate the sentiment of each tweet and classify it in one of the following classes:
\begin{itemize}
    \item \textit{Positive}: Tweets which express happiness, praise a person, group, country or a product, or applaud something.
    \item \textit{Negative}: Tweets which attack a person, group, product or country, express disgust or unhappiness towards something, or criticise something.
    \item \textit{Neutral}: Tweets which state facts, give news or are advertisements. In general, those which don't fall into the above 2 categories.
    \item \textit{Indeterminate}: Tweets where it is not easy to assess sentiment or sentiments of both polarities of approximately the same strength exist. Tweets annotated with the indeterminate label were discarded from our analysis.
\end{itemize}

For each set of tweets three native speakers were assigned as annotators. Initially, 100 tweets were sampled for each language and were given to each group of annotators. The annotators were advised to consider only information available in the text, e.g., to not follow links present, and in cases where a tweet includes only news titles to assess the sentiment of the news being shared.

Table \ref{table:annotation_agreement} displays the  inter-annotation agreement based on Cohen's Kappa \cite{cohen1960coefficient}. It is observable that for all three language sets the agreement between annotators is satisfactory with the lowest score, 0.69, being in the Spanish set when considering entries labelled as `Indeterminate' too. It is also worth noting that the divergence between positive and negative labels (which could be the most problematic in our subsequent analysis) was extremely low. Only 9\% (Greece), 3\% (Spain) and 7\% (UK) of all annotated tweets had a contrasting positive/negative or negative/positive labels between any annotator pair. Finally, in order to consolidate the annotations, the final label of each tweet was decided by using the two annotators agreement in each group and in cases of differences the third annotator was used as a tiebreaker. 

\begin{table}[ht]
\begin{adjustbox}{width=\columnwidth,center}
\begin{tabular}{l|c|l|c|l|c|l|}
\cline{2-7}
                                                & \multicolumn{2}{c|}{UK} & \multicolumn{2}{c|}{Spain} & \multicolumn{2}{c|}{Greece} \\ \hline
\multicolumn{1}{|l|}{Including Indeterminate entries}      & \multicolumn{2}{c|}{0.72}    & \multicolumn{2}{c|}{0.69}    & \multicolumn{2}{c|}{0.72}  \\ \hline
\multicolumn{1}{|l|}{Excluding Indeterminate entries} & \multicolumn{2}{c|}{0.74}    & \multicolumn{2}{c|}{0.75}    & \multicolumn{2}{c|}{0.76}  \\ \hline
\end{tabular}
\end{adjustbox}
\caption{Inter-annotator agreement based on Cohen's Kappa}
\label{table:annotation_agreement}
\end{table}

Having establish an acceptable agreement between the annotators each one was given 300 extra tweets to label. In total, 964, 936 and 965 tweets were collected and labelled for English, Spanish and Greek respectively (the final numbers slightly vary given the different number of discarded tweets with the indeterminate label). The label distribution of the annotated data appears to be different for each language (Table \ref{table:annotation_distribution}) with negative tweets being the most dominant class in Spanish and Greek, in contrast to English where most tweets were labelled as positive.

\begin{table}[ht]
\begin{adjustbox}{center}
\begin{tabular}{|l|c|c|c|}
\hline
\textbf{Label \textbackslash Country} & \textbf{UK} & \textbf{Spain} & \textbf{Greece} \\ \hline
Positive                              & 471         & 366            & 275             \\ \hline
Neutral                               & 240         & 155            & 326             \\ \hline
Negative                              & 253         & 415            & 364             \\ \hline
Indeterminate                         & 36          & 64             & 35              \\ \hline
\end{tabular}
\end{adjustbox}
\caption{Distribution of labels in the annotated dataset for each country.}
\label{table:annotation_distribution}
\end{table}

\subsection{Experimental Setting} \label{sec:expsetting}
\subsubsection{Data} For the training purposes six different datasets are sourced. For each language their respective set of annotated tweets, \textit{Annotated Set} (\textit{in-domain}, see Section \ref{sec:annotation}), is being utilised along with another language specific dataset for each language (\textit{out-domain}): the sentiment analysis dataset from `SemEval-2017 Task 4' \cite{rosenthal2017semeval} is acquired for English (\textit{out-domain} for English); the InterTASS corpus \cite{diaz2018democratization} for Spanish (\textit{out-domain} for Spanish); and a sentiment dataset constituted of tweets related to the 2015 Greek elections for Greek (\textit{out-domain} for Greek) \cite{tsakalidis2018building}. These additional sources are used only for training purposes. All the datasets  have been constructed for the specific task of Twitter sentiment analysis where each tweet is classified as either Positive, Negative, or Neutral. All Twitter handles are anonymised by replacing them by `@username'.

\subsubsection{Training/Evaluation} We consider two different approaches to evaluate our models. Firstly, a train/validation/test split method is applied. The testing data are the subset of tweets from \textit{Annotated Set} that are cross-annotated by all annotators (approximately 100 tweets for each language), the rest of the entries (approximately 900 tweets per language) are used for training and validation with a 85/15 train/validation ratio. 
The cross-annotated subsets are used for testing as it is assumed to be more precise than the larger subset annotated by a single person.

Secondly, due to the relatively small size of the \textit{Annotated Set}, a 5-fold cross-validation method is also applied where the whole dataset is used. In cases where a multilingual model is trained, the combined \textit{Annotated Set} is used along with a stratification method that ensures that all languages are equally represented in each fold. 
The stratification process was based on existing functionality provided by scikit-learn\footnote{\url{https://scikit-learn.org/stable/modules/generated/sklearn.model_selection.StratifiedKFold.html}} and aims to preserve the percentage of tweets of each country and not necessary of each sentiment in every fold.
This cross-validation experiment is set to complement the evaluation done in the single train/test split, which may be limiting \cite{gorman-bedrick-2019-need}.

\subsubsection{Comparison systems} Multiple transformer-based language models, including domain-specific models which are state of the art in Twitter-related classification tasks \cite{barbieri2020tweeteval} are selected for our experiment.
\begin{enumerate}
    \item XLM-R \cite{conneau2019unsupervised}, is pre-trained on a large multilingual corpus containing 100 languages (CommonCrawl \cite{wenzek2020ccnet}).
    \item XLM-T \cite{barbieri2021xlm} further trained on 198 million multilingual tweets.
    \item XLM-T-Sent \cite{barbieri2021xlm}, similar training to XLM-T but further finetuned for multilingual sentiment analysis.
    \item Three different RoBERTa-Base implementations for each language studied: roberta-base \cite{liu2019roberta},  roberta-base-bne \cite{gutierrezfandino2021spanish} and palobert-base-greek-uncased-v1 \cite{info12080331} for English, Spanish and Greek respectively.
    \item TweetEval-Sent \cite{barbieri2020tweeteval} trained on 58 million English tweets and finetuned for sentiment analysis with the English SemEval-2017 dataset \cite{rosenthal2017semeval}.
    \item Bertweet-Sent, based on the Bertweet \cite{nguyen2020bertweet} implementation and further finetuned for sentiment analysis.
\end{enumerate}

All the models are based on the implementations of the uncased versions provided by Hugging Face \cite{wolf-etal-2020-transformers}, and further finetuned and tested for each language individually, as well as in a multilingual setting using the data collected.

In order to assess the difference between these recent transformer models with more traditional approaches, three baseline models were tested: (1) an SVM model using a combination of frequency features, TF-IDF, and semantic, based on the average of word embeddings  (2) a neural network with two LSTM layers while also utilising pretrained word embeddings\footnote{As pre-trained word embeddings, we used 300-dimensional fasttext embeddings \cite{grave2018learning} trained on Wikipedia data} and (3) a lexicon based approach by utilising VADER \cite{hutto2014vader} \footnote{Spanish and Greek texts were first translated to English with Google's Translate API prior to using VADER}.

\subsubsection{Optimisation} All models were trained with the same set of hyper parameters. Specifically, Adam optimiser \cite{loshchilov2017decoupled} and a linear scheduler with warm-up are applied. We warm up linearly for 50 steps with a learning rate of 5e-5, while a batch size n=16 is used. The models are trained up to 20 epochs, with a checkpoint in every epoch, while an early-stop callback stops the training process after 3 epochs without a performance increase of at least 0.01. We select the best model out of all the checkpoints based on their performance on the validation set.

As our aim was not to necessarily find the best sentiment classifier but instead to acquire a robust classifier with good overall performance across all the languages studied no hyper-parameter tuning was performed.

\subsubsection{Evaluation metrics} 

We report results both in the usual macro-average F1 and the F1 average between positive and negative classes (F1\textsuperscript{PN} henceforth). For sentiment analysis tasks the average of the F1 scores of the Positive and Negative classes is often used as an evaluation measure \cite{rosenthal2019semeval} instead of other metrics such as Accuracy. This is mainly justified as firstly F1 scores are more robust to class imbalance, and secondly due to the fact that classifying correctly Positive and Negative classes is more crucial than the Neutral class, especially in our subsequent analysis.

\subsection{Results} 

Table \ref{table:f1_results} displays the average F1 score for only the Positive and Negative classes, F1\textsuperscript{PN}, of the models trained in 5-fold cross-validation experimental setting (More detailed results can be found in Table \ref{tab:f1_averages} in the Appendix).
The performance of the classifiers varies depending not only in their architecture but also on the data that they are trained on. In the UK dataset, the default implementation of `Bertweet-Sent' outperforms all the other models achieving a F1\textsuperscript{PN} of 83\%. For Spanish tweets the multilingual version of `XLM-T-Sent' performs the best with F1\textsuperscript{PN} cross-validation score of 81\%. Finally, when considering the Greek dataset the results are not as clear as in a cross-validation setting the multilingual implementation of `XLM-R' seems to perform better (F1\textsuperscript{PN}=78\%) while in the train/test split setting the implementation of `XLM-R' trained only on both Greek datasets (\textit{out-domain} and \textit{Annotated Set}) performs the best while achieving a similar score.

\subsection{Model selection}
\label{sec:modelselection}
Considering the focus of our study, classifying tweets from members of parliaments in different countries, we decided against the use of mono-lingual models such as `Bertweet-Sent'\footnote{In Section \ref{sec:sentiment_analysis_models} we further present a control analysis in which we compare the trends between our selected multilingual model and the best performing `Bertweet-Sent' in English.} for two reasons: (1) there is no certainty that a tweet will follow the same language as the main language of the parliament, e.g., Welsh tweets in the UK parliament, Catalan tweets in the Spanish parliament, Turkish tweets in the Greek parliament; and (2) using a multilingual model will make the comparison across countries easier.  As such, for the purposes of our experiment, the multilingual implementation of \textbf{`XLM-T-Sent' fine-tuned on our in-domain data} is selected as a classifier and applied across all of the data collected.\footnote{This XLM-T-Sent model is publicly available at \url{https://huggingface.co/cardiffnlp/xlm-twitter-politics-sentiment.}} Our choice is further justified as the selected option produces consistently strong results in all countries (73\%, 87\%, 76\%  F1\textsuperscript{PN} score for UK, Spain and Greece, respectively, when considering the train/test split setting), including state-of-the-art results for Spanish and Greek. 

\begin{table}[H]
\begin{tabular}{|ll|c|c|c|}
\hline
\multicolumn{1}{|l|}{\textbf{\begin{tabular}[c]{@{}l@{}}Train\\ setting\end{tabular}}} &
  \textbf{Model} &
  \textbf{UK} &
  \textbf{Spain} &
  \textbf{Greece} \\ \hline
\multicolumn{1}{|l|}{\multirow{9}{*}{\rot{Monolingual}}}  & XLM-T          & 80          & 80          & 73          \\ \cline{2-5} 
\multicolumn{1}{|l|}{}                              & XLM-T-Sent     & 80          & \textbf{81} & \textbf{75} \\ \cline{2-5} 
\multicolumn{1}{|l|}{}                              & XLM-R          & 74          & 80          & 73          \\ \cline{2-5} 
\multicolumn{1}{|l|}{}                              & RoBERTa-Base   & 80          & 80          & 49          \\ \cline{2-5} 
\multicolumn{1}{|l|}{}                              & TweetEval-Sent & 81          & -           & -           \\ \cline{2-5} 
\multicolumn{1}{|l|}{}                              & Bertweet-Sent  & \textbf{82} & -           & -           \\ \cline{2-5} 
\multicolumn{1}{|l|}{}                              & SVM            & 61          & 58          & 44          \\ \cline{2-5} 
\multicolumn{1}{|l|}{}                              & LSTM           & 67          & 60          & 57          \\ \cline{2-5} 
\multicolumn{1}{|l|}{}                              & VADER          & 64          & 63          & 58          \\ \hline
\multicolumn{1}{|l|}{\multirow{3}{*}{\rot{Multilingual}}} & XLM-T          & 81          & 80          & 75          \\ \cline{2-5} 
\multicolumn{1}{|l|}{}                              & XLM-T-Sent     & 80          & \textbf{81} & \textbf{77} \\ \cline{2-5} 
\multicolumn{1}{|l|}{}                              & XLM-R          & 78          & 76          & 76          \\ \hline
\multicolumn{2}{|l|}{Random - Baseline}                              & 33          & 30          & 27          \\ \hline
\end{tabular}
\caption{Average F1 scores of positive and negative classes (F1\textsuperscript{PN}) when trained/evaluated with 5-fold cross-validation. The results displayed are the averages of three runs. Models were trained on both `in' and `out' of domain . Both multilingual and monolingual train-setting results are reported}
\label{table:f1_results}
\end{table}

\paragraph{Robustness across languages:} As a final check we aimed to evaluate `XLM-T-Sent' performance on languages other than English, Spanish and Greek. To this end, we utilised our Spanish test set which contains 51 entries in Catalan (5.4\% of the whole set) and considered the model’s performance on these entries. Our results (5-fold CV) indicate that the performance remains satisfactory with an overall Accuracy of 86.6\% and F1\textsuperscript{PN} score of 88.9\% when considering only the Catalan tweets. For reference, when looking only at the Spanish tweets in the test set, 'XLM-T-Sent' achieves 87.4\% and 90\% Accuracy and F1\textsuperscript{PN} scores respectively.

The above results, along with the fact that `XLM-T-Sent' has originally been trained in a total of 100 languages including Catalan, Basque and Welsh \cite{conneau2019unsupervised}, and previous work indicating XLM-R’s cross-lingual transfer learning capabilities \cite{barbieri2021xlm, li2021improving, rusliapplicability}, provide further reassurances that our classifier is capable on handling the majority of tweets collected to satisfying degree.

\paragraph{Discussion} Overall, our results show that language models perform consistently better than traditional machine learning approaches in our setting. Even though our experiments do not necessarily provide a clear answer to RQ1 they do help us acquire a stable and robust sentiment classifier that can be used across multiple languages. The selected multilingual model (i.e., `XLM-T-Sent' which has been further fine-tuned on our labelled in-domain data) is then utilized for the rest of the experiments, and in particular for attempting to answer research questions RQ2 and RQ3.

\section{Analysis}\label{sec:analysis}

Having acquired a suitable sentiment analysis classifier capable to successfully distinguish sentiment polarity in MP tweets (see Section \ref{sec:modelselection}) we applied it to our collected Twitter corpora (see Section \ref{sec:datacollect}) and perform an in-depth analysis aiming to explore whether politicians' tweets containing negative sentiment have a bigger network penetration than tweets that are positive or neutral.

Initially, we attempt to establish what is considered a `popular' tweet in the context of our analysis. Table \ref{tab:tweets_percentiles} displays the percentiles of how many times a tweet has been retweeted for the UK, Spanish and Greek parliaments (2021). It is noticeable that the vast majority of tweets are retweeted only a few times; 75\% of tweets having a retweet count below 40 across all parliaments, indicating a long tail in the distribution of the retweets count. In our analysis, we consider a tweet to be `popular' and to belong in the \textbf{`Head'} of the distribution when it is included in the top 5\textsuperscript{th} percentile of the retweets count. On the other hand, a tweet is labelled as `not popular' and it belongs in the \textbf{`Tail'} of the distribution when it falls under the 50\textsuperscript{th} percentile.

\begin{table}[hbt!]
\begin{adjustbox}{width=\columnwidth,center}
\begin{tabular}{|l|c|c|c|c|c|c|}
\hline
Parliament\textbackslash{}Pct & 0\% & 25\% & \textbf{50\%} & 75\% &  \textbf{95\%} & 100\% \\ \hline
UK & 0 & 0 & 2 & 9 & 52 & 36689 \\ \hline
Spanish & 0 & 1 & 5 & 36 & 185 & 25132 \\ \hline
Greek & 0 & 0 & 3 & 9 & 28 & 3357 \\ \hline
\end{tabular}
\end{adjustbox}
\caption{Percentiles of retweet counts for UK, Spanish and Greek parliaments from January to December 2021. Highlighted are the percentiles used for determining if a tweet belongs in the `Head' or `Tail'.}
\label{tab:tweets_percentiles}
\end{table}

\subsection{Sentiment and Virality}
\label{sec:sentvir}

To directly answer the research question RQ2, we investigate whether there exists a correlation between retweets count and sentiment based on our collected data (see Section \ref{sec:datacollect}) and sentiment analysis classifiers (see Section \ref{sec:models_evaluation}).

\subsubsection{Correlation analysis}

As an initial step, we ran several correlation experiments based on statistical testing and regression analysis. For the purposes of these experiments, only tweets from the \textit{2021 Main Dataset} (see Section \ref{maindataset}) were utilised.

\paragraph{Statistical Testing} As the comparison includes numerical (retweets count) and categorical values (sentiment) approaches such as the Spearman's correlation are not suited. Instead we perform a Chi-Square test which indicates the existence of a dependency between popularity and sentiment. Then the Kruskal-Wallis H-test is performed on the retweets count populations for Positive, Negative and Neutral charged tweets to test whether their median values differ. Our test clearly confirms the existence of a difference in the distributions of the populations among sentiment labels (\textit{p-value} $<10^{-16}$, \textit{a}=$0.05$)\footnote{\textit{p-value} is lower than the minimum accepted value in Python.}. Even though there is no evidence for direct correlation, we manage to establish that there is a relation between sentiment and popularity, and also that retweet count distributions differ between sentiment labels.

\paragraph{Regression Analysis} To complement the statistical tests multiple regression models are fitted and the significance of sentiment is examined.\footnote{All regression models were based on the implementations of Python's `statsmodels' library  \cite{seabold2010statsmodels}.}. In our experimental setting, the retweet count is set as the dependent variable while the existence (or not) of positive and negative sentiment constitutes the independent variables. Neutral sentiment is not taken into account in order to avoid potential collinearity problems. Furthermore, we utilise four lexical statistics (Table \ref{tab:control_stats}) as control variables. Specifically, the presence of emojis, URLs, hashtags (\#),  and mentions (@) in a tweet are used as binary variables \footnote{The percentage of uppercase characters was not used in order to avoid influencing the rest of the binary predictors.}. We did not consider features such as number of followers, favourites count or time posted as our main focus was to identify the importance of features extracted from text.

Due to the nature of the problem at hand, i.e., modelling a count variable, and to the highly skewed distribution of the target (see Table \ref{tab:tweets_percentiles}) we test a negative binomial regression model (\textit{NBR}) \cite{hilbe2011negative} and a zero inflated variation of it (\textit{zero-NBR}). Previous research suggests that \textit{NBR} models and the zero inflated variations can be successfully used to estimate tweets popularity (i.e., retweet count) \cite{bhattacharya2014engagement, sutton2015cross, samuel2020message}. \textit{NBR} is a generalised linear model used for modelling count data where the dependent variable is assumed to follow the negative binomial distribution. In contrast to Poisson regression, which is also used to model count variables, \textit{NBR} does not make the assumption that the mean is equal to variance making it more robust when dealing with over-dispersed data. \footnote{The zero inflated variations are based on the assumption that the dependent variable's distribution displays an excess of zero-valued observations and combines two processes: one process (logistic regression) that generates zeros and a second one that generates counts according to the negative binomial model or Poisson model.}

For our setting we train six different \textit{NBR} models using both the combined dataset of the UKs, Spanish and Greek parliaments and data from each individual parliament. The results of both \textit{NBR} and \textit{zero-NBR} models (Table \ref{tab:binomial_reg}) indicate that both negative and positive sentiment are statistically significant in predicting the retweet count of a tweet. At the same time, when considering the coefficients of the two variables we can observe that negative sentiment is assigned a greater weight than the positive sentiment with both models assigning  coefficient over 5 times larger to negative sentiment when trained on all three parliaments. In fact, for the UK parliament positive sentiment has even an overall negative impact on the retweet count. Finally, it is interesting to note that even though every feature tested appears to be statistically significant, the negative sentiment appears to be the most effective on spreading a tweet. 

\begin{table*}[ht]
\begin{adjustbox}{width=\textwidth,center}
\begin{tabular}{|l|cc|cc|cc|cc|}
\hline
\multirow{2}{*}{\textbf{variable}} &
  \multicolumn{2}{c|}{\textbf{Overall}} &
  \multicolumn{2}{c|}{\textbf{UK}} &
  \multicolumn{2}{c|}{\textbf{Spain}} &
  \multicolumn{2}{c|}{\textbf{Greece}} \\ \cline{2-9} 
 &
  \multicolumn{1}{c|}{\textbf{NBR}} &
  \textbf{zero NBR} &
  \multicolumn{1}{c|}{\textbf{NBR}} &
  \textbf{zero NBR} &
  \multicolumn{1}{c|}{\textbf{NBR}} &
  \textbf{zero NBR} &
  \multicolumn{1}{c|}{\textbf{NBR}} &
  \textbf{zero NBR} \\ \hline
const &
  \multicolumn{1}{c|}{3.0756*} &
  3.4951* &
  \multicolumn{1}{c|}{3.2976*} &
  3.7691* &
  \multicolumn{1}{c|}{3.373*} &
  3.6191* &
  \multicolumn{1}{c|}{2.2644*} &
  2.3446* \\ \hline
emojis &
  \multicolumn{1}{c|}{-0.1868*} &
  -0.1886* &
  \multicolumn{1}{c|}{-0.2776*} &
  -0.3187* &
  \multicolumn{1}{c|}{-0.2431*} &
  -0.1643* &
  \multicolumn{1}{c|}{-0.2524*} &
  -0.2535* \\ \hline
has\_url &
  \multicolumn{1}{c|}{0.4031*} &
  -0.0417* &
  \multicolumn{1}{c|}{0.0864*} &
  -0.4065* &
  \multicolumn{1}{c|}{0.8029*} &
  0.5433* &
  \multicolumn{1}{c|}{-0.1167*} &
  -0.2282* \\ \hline
hashtags &
  \multicolumn{1}{c|}{-0.3745*} &
  -0.5002* &
  \multicolumn{1}{c|}{-0.2173*} &
  -0.3105* &
  \multicolumn{1}{c|}{-0.402*} &
  -0.442* &
  \multicolumn{1}{c|}{-0.2642*} &
  -0.2833* \\ \hline
mentions &
  \multicolumn{1}{c|}{-0.9496*} &
  -0.6461* &
  \multicolumn{1}{c|}{-1.1553*} &
  -0.6113* &
  \multicolumn{1}{c|}{-1.0098*} &
  -0.8653* &
  \multicolumn{1}{c|}{-0.7287*} &
  -0.6088* \\ \hline
neg &
  \multicolumn{1}{c|}{1.4476*} &
  1.3516* &
  \multicolumn{1}{c|}{1.3666*} &
  1.231* &
  \multicolumn{1}{c|}{1.1611*} &
  1.1297* &
  \multicolumn{1}{c|}{0.8549*} &
  0.8589* \\ \hline
pos &
  \multicolumn{1}{c|}{0.2766*} &
  0.2438* &
  \multicolumn{1}{c|}{-0.0796*} &
  -0.1364* &
  \multicolumn{1}{c|}{0.4683*} &
  0.3971* &
  \multicolumn{1}{c|}{0.3472*} &
  0.3525* \\ \hline
\end{tabular}
\end{adjustbox}
\caption{Coefficients for the Negative Binomial Regression (NBM) and zero inflated NBM. $*$ indicates p-value $< 0.05$ }
\label{tab:binomial_reg}
\end{table*}

To verify our regression results we also test Poisson regression \cite{coxe2009analysis} and zero inflated Poisson regression models. Poisson regression is also a generalised linear model that assumes that dependent variable follows the Poisson distribution. Similar to \textit{NBR} it is used to model count data but makes stricter assumptions ($mean = variance$). Both implementations confirm our original results indicating that while both negative and positive sentiments are significant, negative sentiment appears to be larger factor on the tweet popularity. Finally, a simple ordinary least square model (\textit{OLS}) is used while using the softmax outputs of \textit{XLM-T-sent} as dependent variable. For each tweet we select the highest softmax score (only negative or positive scores are considered) and assign negative and positive signs to each sentiment respectively. By combining the two variables we acquire an overview on the overall impact of sentiment and model confidence (not only the final labels) on the tweets popularity. Similar to the previous experiments the \textit{OLS} model identifies the sentiment score as significant and assigns a negative weight implying that the more negative a tweet is the more likely is to be shared. Detailed results of all the models tested can be found in the Appendix (Tables \ref{tab:poisson_results} \& \ref{tab:ols_results}).

To sum up, the results of the statistical tests and regression analysis show a strong correlation between negative sentiment and virality, which help us to answer RQ2. Specifically, model's coefficients show that the existence of negative sentiment in a tweet has a major positive impact on its popularity independently of the country from where it was posted. 

\subsubsection{Head and tail distribution analysis}

\begin{figure}[H]
\centering
\begin{adjustbox}{width=\columnwidth,center}
\includegraphics{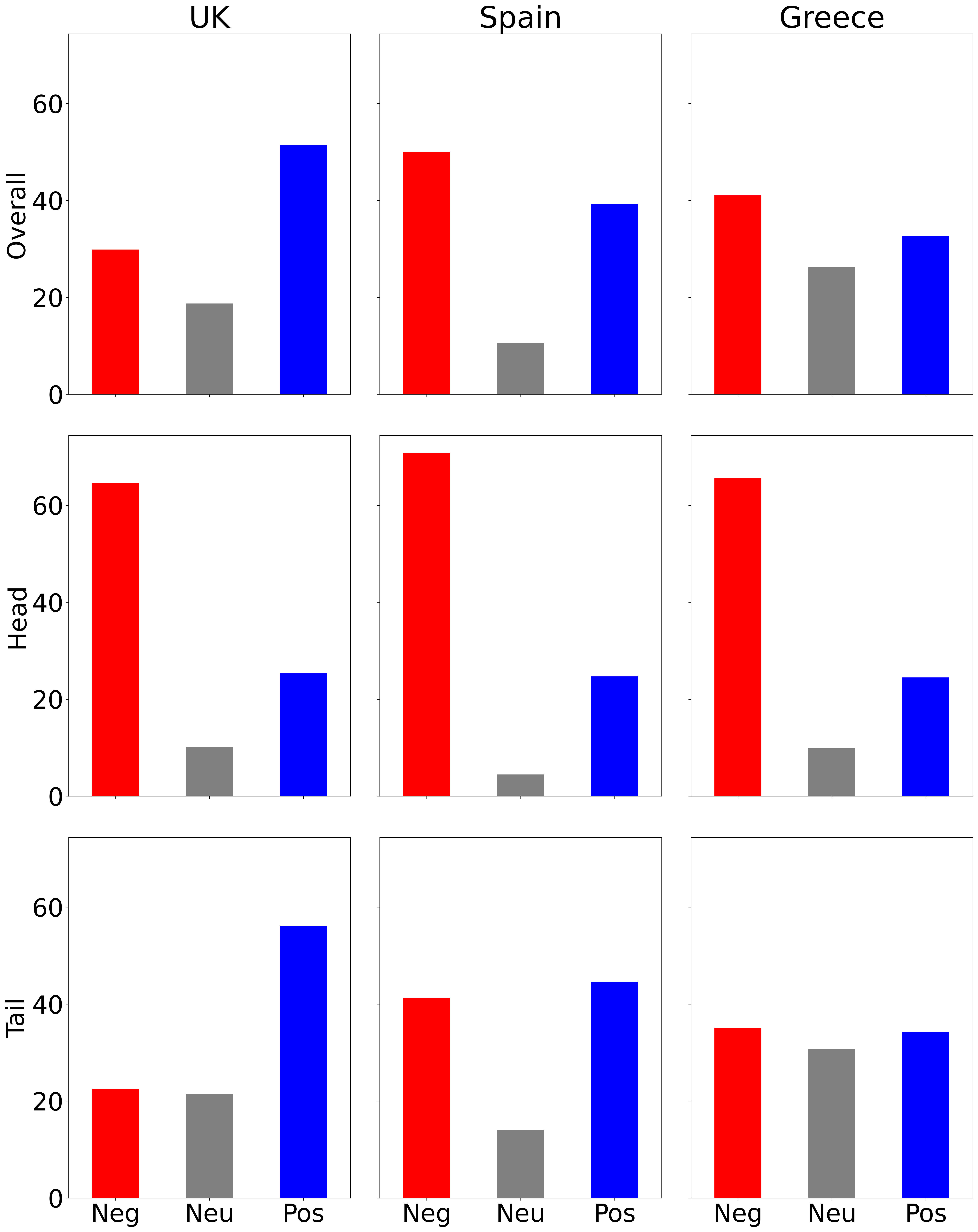}
\end{adjustbox}
\caption{Sentiment distribution (Overall, Head, Tail) for tweets from the UK, Spanish, Greek parliaments in 2021. From left to right: Negative sentiment (red), Neutral sentiment (grey), Positive sentiment (blue).}
\label{fig:2021_sentiment_distribution}
\end{figure}

For this extended analysis, we consider the tweets made from politicians of the main parliaments of the UK, Spain, and Greece separately, and compare them based on their popularity following the `Head' and `Tail' partitions presented at the beginning of the section. Figure \ref{fig:2021_sentiment_distribution} displays the normalised counts of positive, negative and neutral tweets in the \textit{2021 Main Dataset}. 
Looking at the overall tweets distribution, negative account for a higher percentage for the Spanish and Greek parliaments whilst the reverse is true for the UK. However, when comparing the most `popular' tweets (`Head') to those having only a small number of retweets count (`Tail'), there is clear pattern with negative charged messages being more numerous for all parliaments.
For the UK parliament when comparing the `Head' and `Tail' sets, the proportion of negative tweets is higher for the most `popular'; `Tail' tweets with negative emotion are 65\% less than those of the `Head', whilst positive tweets display a 121\% increase between `Head' and `Tail'. Similarly, negative tweets are 71\% and 88\% more numerous in the Spanish and Greek parliaments, respectively, when comparing most to least popular tweets.

\subsection{Governing Party vs Opposition}
\label{sec:opposition}

\begin{figure}[ht]
\centering
\includegraphics[width= 1\linewidth]{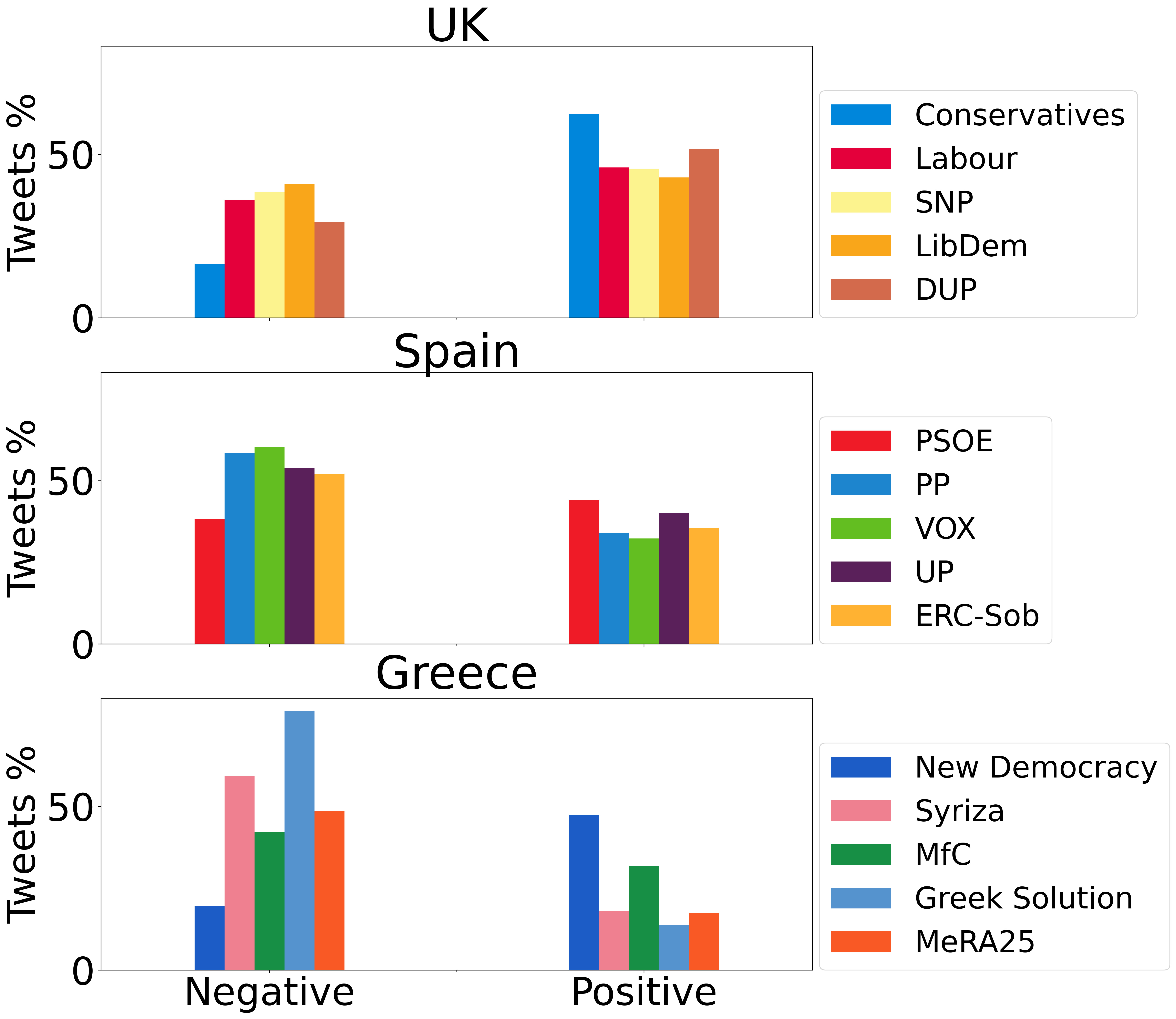}
\caption{Sentiment per party for tweets from the UK, Spanish, Greek parliaments 2021. Top five parties displayed, ordered from left to right by their seat number. }
\label{fig:2021_parties}
\end{figure}

To answer the first part of the researcher question RQ3 we consider the sentiment distribution on the party level for each parliament. Figure \ref{fig:2021_parties} presents the difference in sentiment for the top five major parties (based on the number of MPs) in the UK's, Spain's and Greece's parliaments for 2021. Taking into account the governing parties of each country (UK: Conservatives, Spain: Spanish Socialist Workers' Party (PSOE) \& Unidas Podemos (UP), Greece: New Democracy) a distinctive trend appears. In each country, the main governing parties tweets tend to be significantly more positive than those of the opposition. In the UK, there are 46\% more positive tweets than negative posted by the ruling party (Conservatives) whereas only a 10\% difference for the main opposition (Labour party). The same pattern appears for Greece with positive charged tweets posted by New Democracy being 28\% more than the negative. At the same time, the opposition party, Syriza, being on the `attack' has 59\% of its total tweets posted classified as negative. Similarly, for Spain, even though to a smaller degree, PSOE display the biggest contrast between positive and negative tweets with only a 6\% difference in favour of positive. 
On a final note, it is interesting to observe that the VOX and Greek Solution, two right-wing parties, display the largest percentages of negative tweets for their respective parliaments (60\% and 79\%).

This behaviour becomes even more apparent when we take into consideration only the tweets from the leaders of the governing and opposition parties (Table \ref{tab:leaders_sentiment}). The UK's Prime Minister (Boris Johnson) shares a considerably larger percentage of positive tweets compared to the opposition leader Keir Starmer (79\% vs. 58\%) and only 5\% of his tweets being negative, in contrast to the 35\% for Keir Starmer. An even bigger contrast is seen between the political leaders in Spain where 84\% of all tweets from President Sánchez are negative, compared to only 30\% from the opposition leader Pablo Casado. In Greece the trend is similar, with a high contrast of positive/negative tweets between the Prime Minister Mitsotakis and the opposition leader Tsipras (77\%/17\% vs. 29\%/59\%). 

\begin{table}[ht]
\begin{adjustbox}{width=0.47\textwidth,center}
\begin{tabular}{|l|l|l|l|l|l|}
\hline
\multicolumn{1}{|c|}{Parliament} &
  \multicolumn{1}{c|}{Party} &
  \multicolumn{1}{c|}{Name} &
  \multicolumn{1}{c|}{Neg} &
  \multicolumn{1}{c|}{Neu} &
  \multicolumn{1}{c|}{Pos} \\ \hline
\multirow{2}{*}{UK}     & Conservatives & Boris Johnson    & 5  & 16 & 79 \\ \cline{2-6} 
     & Labour        & Keir Starmer    & 35 & 7  & 58 \\ \hline
\multirow{2}{*}{Spain}  & PSOE          & Pedro Sánchez  & 13 & 3  & 84 \\ \cline{2-6} 
  & PP            & Pablo Casado Blanco  & 63 & 6  & 30 \\ \hline
\multirow{2}{*}{Greece} & ND & Kyriakos Mitsotakis   & 17 & 6  & 77 \\ \cline{2-6} 
 & Syriza        & Alexis Tsipras  & 59 & 12 & 29 \\ \hline
\end{tabular}
\end{adjustbox}
\caption{Sentiment distribution for government leaders  (Johnson, Sanchez, Mitsotakis) and opposition (Starmer, Blanco, Tsipras) leaders.}
\label{tab:leaders_sentiment}
\end{table}

Therefore, these observations support the hypothesis that politicians use Twitter to promote their agenda. 
Not surprisingly, the governing parties try to depict a positive image of the state of their country. In contrast, the opposition parties challenge the same positions by using negatively charged tweets.

\subsection{Devolved Parliaments}
\label{sec:devolved}

Aiming to acquire a more detailed representation for each country, as well as to answer the second part of the research question RQ3, we applied our sentiment analysis pipeline to the devolved parliaments, \textit{2021 Devolved Dataset}, of the UK (N.Ireland, Scotland, Wales) and Spain (Catalonia, Basque country). Table \ref{tab:sentiment_dist} displays the sentiment distribution (Overall, `Head' and `Tail') in the main and devolved parliaments for each country. The devolved parliaments of the UK follow a similar pattern to its main parliament, with more positive than negative tweets overall. In contrast, for the Spanish parliaments there is no consistent pattern with the Basque and main parliaments being dominated by negatively charged tweets whereas in the Catalan parliament the tweets tend to be more positive inclined.

\begin{table}[ht]
\begin{adjustbox}{width=0.47\textwidth,center}
\centering

\begin{tabular}{|c|c|c|cc|cc|cc|}
\hline
\multirow{2}{*}{\textbf{Source}} &
  \multirow{2}{*}{\textbf{Year}} &
  \multirow{2}{*}{\textbf{Country}} &
  \multicolumn{2}{c|}{\textbf{Overall}} &
  \multicolumn{2}{c|}{\textbf{Head}} &
  \multicolumn{2}{c|}{\textbf{Tail}} \\ \cline{4-9} 
 &
   &
   &
  \multicolumn{1}{c|}{Neg} &
  Pos &
  \multicolumn{1}{c|}{Neg} &
  Pos &
  \multicolumn{1}{c|}{Neg} &
  Pos \\ \hline
\multirow{10}{*}{Parliament} &
  2014 &
  \multirow{3}{*}{UK} &
  \multicolumn{1}{c|}{\ccn{29}} &
  \ccp{41} &
  \multicolumn{1}{c|}{\ccn{43}} &
  \ccp{24} &
  \multicolumn{1}{c|}{\ccn{27}} &
  \ccp{46} \\ \cline{2-2} \cline{4-9} 
 &
  2016 &
   &
  \multicolumn{1}{c|}{\ccn{29}} &
  \ccp{46} &
  \multicolumn{1}{c|}{\ccn{48}} &
  \ccp{31} &
  \multicolumn{1}{c|}{\ccn{27}} &
  \ccp{47} \\ \cline{2-2} \cline{4-9} 
 &
  \multirow{14}{*}{2021} &
   &
  \multicolumn{1}{c|}{\ccn{30}} &
  \ccp{51} &
  \multicolumn{1}{c|}{\ccn{65}} &
  \ccp{25} &
  \multicolumn{1}{c|}{\ccn{22}} &
  \ccp{56} \\ \cline{3-9} 
 &
   &
  -N.Ireland &
  \multicolumn{1}{c|}{\ccn{36}} &
  \ccp{44} &
  \multicolumn{1}{c|}{\ccn{48}} &
  \ccp{34} &
  \multicolumn{1}{c|}{\ccn{39}} &
  \ccp{41} \\ \cline{3-9} 
 &
   &
  -Scotland &
  \multicolumn{1}{c|}{\ccn{31}} &
  \ccp{53} &
  \multicolumn{1}{c|}{\ccn{60}} &
  \ccp{27} &
  \multicolumn{1}{c|}{\ccn{26}} &
  \ccp{57} \\ \cline{3-9} 
 &
   &
  -Wales &
  \multicolumn{1}{c|}{\ccn{22}} &
  \ccp{60} &
  \multicolumn{1}{c|}{\ccn{37}} &
  \ccp{49} &
  \multicolumn{1}{c|}{\ccn{23}} &
  \ccp{60} \\ \cline{3-9} 
 &
   &
  Spain &
  \multicolumn{1}{c|}{\ccn{50}} &
  \ccp{39} &
  \multicolumn{1}{c|}{\ccn{71}} &
  \ccp{25} &
  \multicolumn{1}{c|}{\ccn{41}} &
  \ccp{45} \\ \cline{3-9} 
 &
   &
  -Catalonia &
  \multicolumn{1}{c|}{\ccn{41}} &
  \ccp{52} &
  \multicolumn{1}{c|}{\ccn{68}} &
  \ccp{28} &
  \multicolumn{1}{c|}{\ccn{34}} &
  \ccp{58} \\ \cline{3-9} 
 &
   &
  -Basque &
  \multicolumn{1}{c|}{\ccn{56}} &
  \ccp{34} &
  \multicolumn{1}{c|}{\ccn{75}} &
  \ccp{22} &
  \multicolumn{1}{c|}{\ccn{54}} &
  \ccp{36} \\ \cline{3-9} 
 &
   &
  Greece &
  \multicolumn{1}{c|}{\ccn{41}} &
  \ccp{33} &
  \multicolumn{1}{c|}{\ccn{66}} &
  \ccp{24} &
  \multicolumn{1}{c|}{\ccn{35}} &
  \ccp{34} \\ \cline{1-1} \cline{3-9} 
\multirow{3}{*}{Random} &
   &
  UK &
  \multicolumn{1}{c|}{\ccn{30}} &
  \ccp{51} &
  \multicolumn{1}{c|}{\ccn{27}} &
  \ccp{54} &
  \multicolumn{1}{c|}{\ccn{31}} &
  \ccp{51} \\ \cline{3-9} 
 &
   &
  Spain &
  \multicolumn{1}{c|}{\ccn{46}} &
  \ccp{39} &
  \multicolumn{1}{c|}{\ccn{43}} &
  \ccp{43} &
  \multicolumn{1}{c|}{\ccn{46}} &
  \ccp{38} \\ \cline{3-9} 
 &
   &
  Greece &
  \multicolumn{1}{c|}{\ccn{25}} &
  \ccp{26} &
  \multicolumn{1}{c|}{\ccn{28}} &
  \ccp{40} &
  \multicolumn{1}{c|}{\ccn{26}} &
  \ccp{25} \\ \cline{1-1} \cline{3-9} 
\multirow{3}{*}{Verified} &
   &
  UK &
  \multicolumn{1}{c|}{\ccn{23}} &
  \ccp{43} &
  \multicolumn{1}{c|}{\ccn{34}} &
  \ccp{38} &
  \multicolumn{1}{c|}{\ccn{21}} &
  \ccp{45} \\ \cline{3-9} 
 &
   &
  Spain &
  \multicolumn{1}{c|}{\ccn{31}} &
  \ccp{39} &
  \multicolumn{1}{c|}{\ccn{45}} &
  \ccp{34} &
  \multicolumn{1}{c|}{\ccn{31}} &
  \ccp{38} \\ \cline{3-9} 
 &
   &
  Greece &
  \multicolumn{1}{c|}{\ccn{38}} &
  \ccp{42} &
  \multicolumn{1}{c|}{\ccn{38}} &
  \ccp{45} &
  \multicolumn{1}{c|}{\ccn{39}} &
  \ccp{41} \\ \hline
\end{tabular}
\end{adjustbox}
 \caption{Overall, Head \& Tail distributions of sentiment for the datasets utilised. Numbers in the table correspond to the percentage of tweets labelled as negative and positive for each subset.}
 \label{tab:sentiment_dist}
 \end{table}

Irrespective of these differences, all the devolved parliaments seem to follow the same general trend where tweets conveying negative sentiment travel further. Similar to their respective main counterparts, we observe that in the `Head' of each devolved parliament negative tweets tend to be more numerous than positive ones. The only exception being the Welsh parliament (Senedd) where positive tweets are the majority (49\% positive to 37\% negative). These findings provide more evidence to the hypothesis of a higher network penetration of negatively charged tweets. It is worth noting that both for the Catalan and Basque parliaments our sentiment model classifies as Neutral only 7\% and 9\% of the total entries which may indicate a higher number of polarised tweets. Moreover, in these regions we can find a more frequent use of less-resourced languages which the sentiment analysis model may find it harder to deal with: Catalan (62\% of all entries) and Basque (27\% of all entries) .

\section{Control Experiments}
\label{sec:control}

In this section, we present four control experiments to test the robustness of our evaluation.

\begin{figure*}[ht]
\centering
\includegraphics[width=\textwidth]{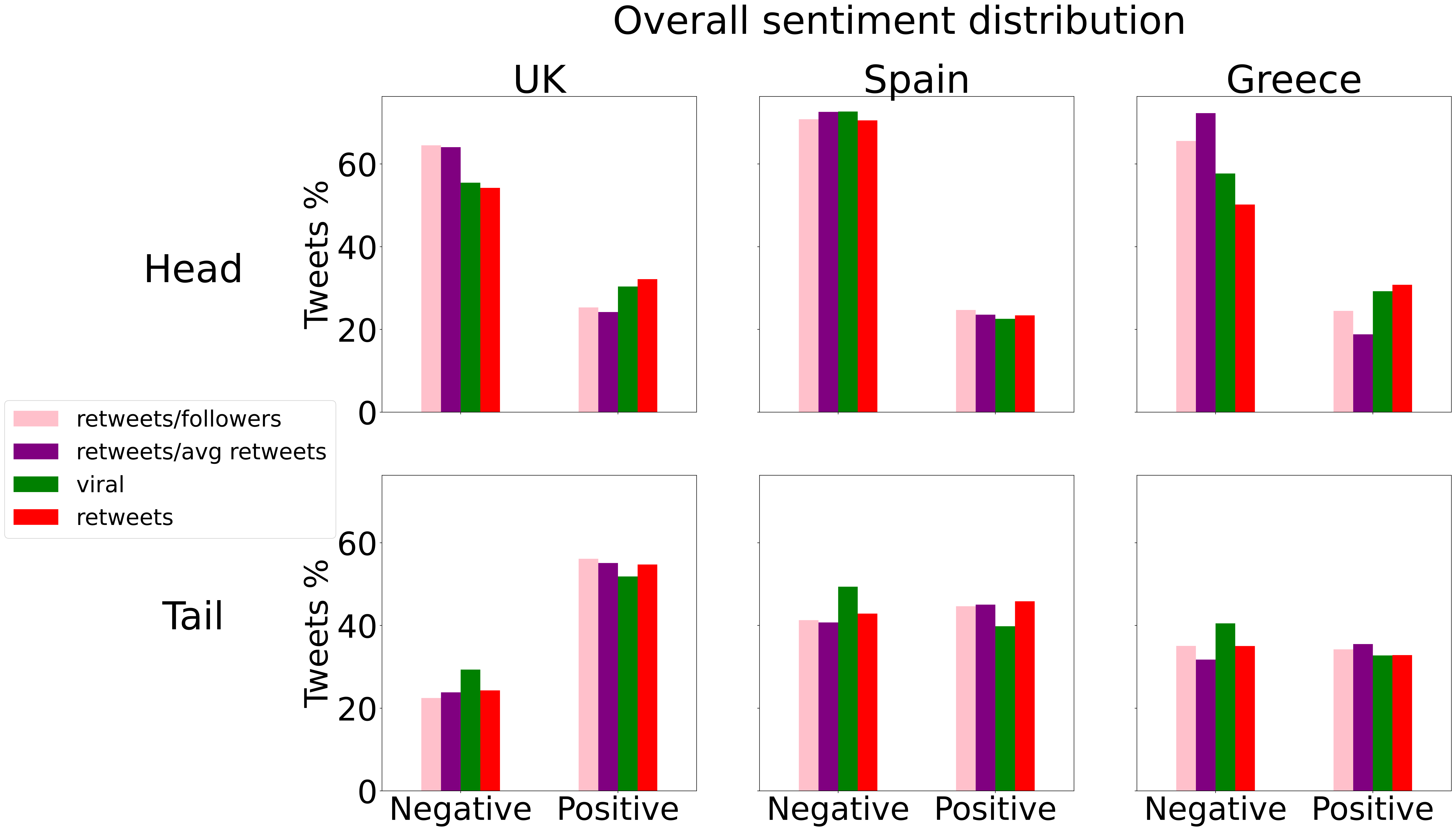}
\caption{Sentiment distribution for tweets from the UK, Spanish, Greek parliaments 2021 based on different definitions of
\textit{tweet popularity}. (1) `number of retweets': pink -- left bar, (2) `number of retweets/ number of followers': purple -- middle bar,  (3) `viral $>= 2 *$ std retweet count`,  (4) `number of retweets/ number of average retweets': green -- right bar.}
\label{fig:2021_sentiment_distribution_other_metrics}
\end{figure*}

\subsection{Popularity metrics}\label{sec:popularity_metrics}

In addition to the raw retweet counts, we also tested additional metrics of popularity to divide the tweets in `Head' and `Tail'. 
As the retweet count is an absolute measure that does not take into account the existing popularity of the user posting a tweet, it may be skewed to favour users with a big number of followers. We attempt to incorporate the popularity of each user and explore whether there are differences in the sentiment trend when using normalised metrics. To achieve this, three new metrics are introduced: (1) the ratio between the retweets count and the follower count of the user; and (2) the ratio of the retweets count to the average number of retweets of the user. This way, a heavily shared tweet from a user that tends to get only few retweets will be considered more `popular' than a similar tweet originating from a user that is retweeted often.  Finally, we also consider a third popularity metric \cite{han2020importance} where a tweet is considered to be viral if its retweet count is at least two standard deviations above the mean for its creator. These three metrics offer an alternative and more normalised view of popularity.

Figure \ref{fig:2021_sentiment_distribution_other_metrics} displays the results of the sentiment distribution for `Head' and `Tail' for the UK's, Spanish and Greek parliaments using the different popularity metrics.
The trends for the four metrics are largely similar, with negative-charged tweets being more popular in all cases. Having established that all four popularity metrics verify the underlying phenomenon, the total retweets count is used as a metric for the rest of our analysis.

\paragraph{Discussion} As there is no clear definition of what a viral tweet is, it is challenging to investigate the effect of sentiment on the popularity of a tweet. By testing multiple popularity metrics we attempt to strengthen our analysis. The results provide further evidence on the effect of negative sentiment on tweet propagation, irrespective of the chosen popularity metric.

\begin{figure}[H]
\centering
\includegraphics[width=\columnwidth]{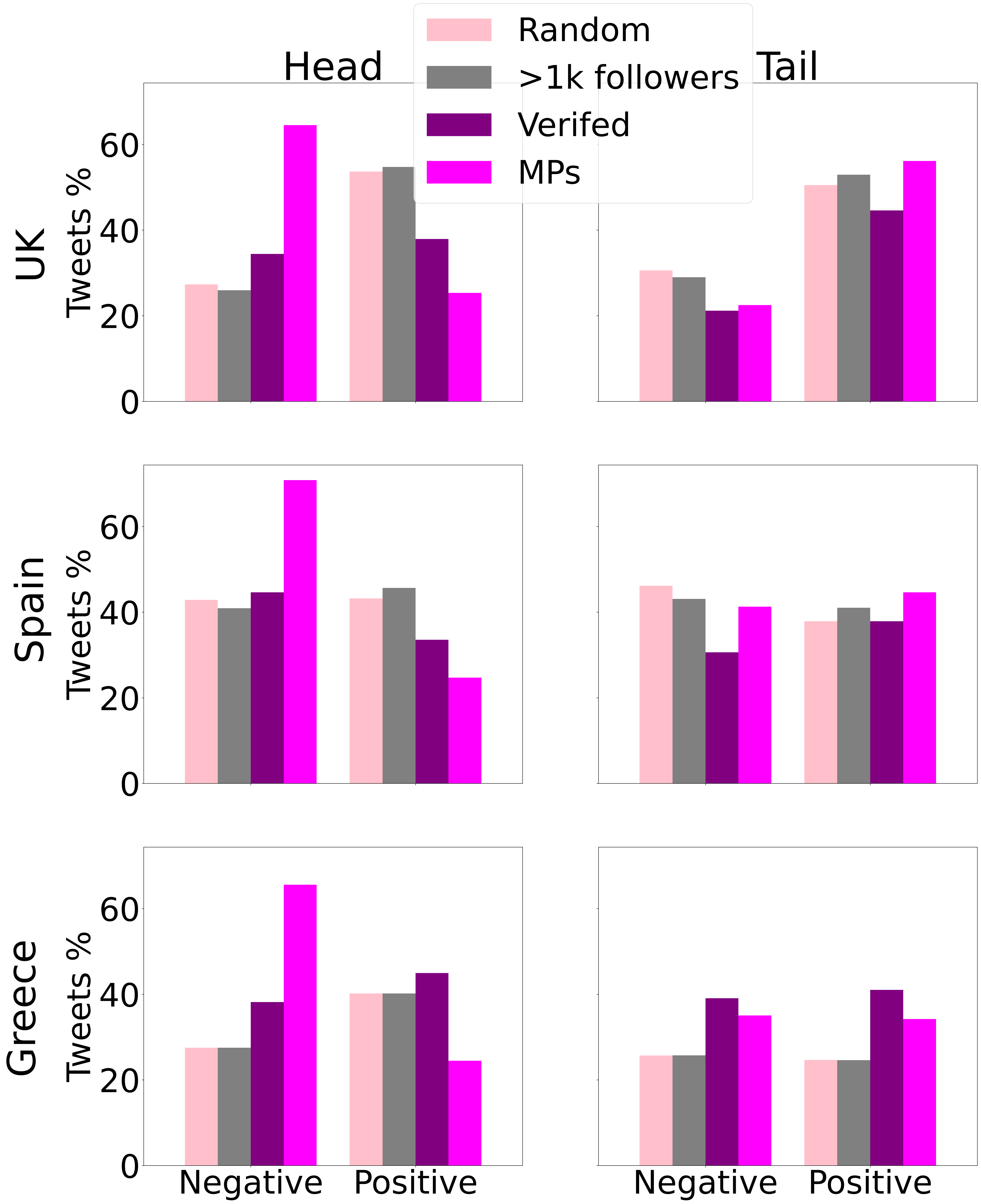}
\caption{Tweets distribution in `Head' and `Tail' based on sentiment for random and verified Twitter users and Members of Parliament. Bars from left to right: (1) Random users: pink, (2) Random users with more than 1000 followers: grey, (3) Verified users: purple, (4) Members of parliaments: fuchsia.}
\label{fig:mps_vs_general}
\end{figure}

\subsection{Politicians Vs General Population}\label{sec:mps_vs_general}

We continue our exploration by investigating whether politicians' tweets spread in the same manner to the general public by comparing the \textit{2021 Main Dataset} with a collection of tweets from random and verified users (\textit{Random \& Verified 2021} dataset, see Section \ref{sec:random&verified} for more details). Figure \ref{fig:mps_vs_general} displays the difference in the sentiment of the tweets amongst these groups for the `Head' and `Tail' of popularity distribution.
In contrast to politicians' tweets, the general population seems to post more positive tweets overall. Moreover, positive tweets are significantly more retweeted in comparison to negative by a big margin. When only considering the most popular tweets (`Head'), in the UK positive tweets for random users are 26\% more numerous than negative, and the same stands for Greece with positive tweets being 13\% more populous while in Spain we observe a small difference in favour of positive (0.3\%). A similar trend occurs when looking at more influential users (random users with more than 1000 followers) where their most shared tweets are mostly positive; 55\%, 46\%, 40\% portion in `Head' for UK, Spain and Greece respectively.

The above results seems to be contradicting the trends observed for MPs. 
Even though politicians' negative tweets are being shared more often it is not the same case for an average/random user. This suggests that users tend to retweet more easily a negatively charged tweet posted by a politician than from another random user.

On the other hand, there is no a clear distinction based on sentiment when considering only verified users among all countries.
UK and Greek verified users tweet positive messages more than negative. However, the opposite phenomenon is observed for Spain  where the proportion of negative charged tweets are significantly higher in the `Head' when the opposite is true when looking at less popular tweets (`Tail'). This could be evidence that Twitter users are more likely to share negatively charged content when it is originating from widely recognisable and influential accounts (artists, athletes, organisations, etc.) or from figures of authority such as politicians, whose negative messages seem to spread faster overall in all countries analysed.

\paragraph{Discussion} This control experiment was aimed at calibrating the predictions made by the fine-tuned sentiment analysis model. By comparing the overall predictions made on our MP datasets with that of the general politicians and verified users, we certified that the trends for sharing positive and negative tweets are indeed different and not widespread Twitter artefacts. In particular, this analysis showed that the proportion of popular negatively charged tweets is higher when coming from an MP in comparison to the rest of the population.

\subsection{Temporal Analysis}\label{sec:temporal_analysis}

\begin{figure}[H]
\centering
\includegraphics[width=\columnwidth]{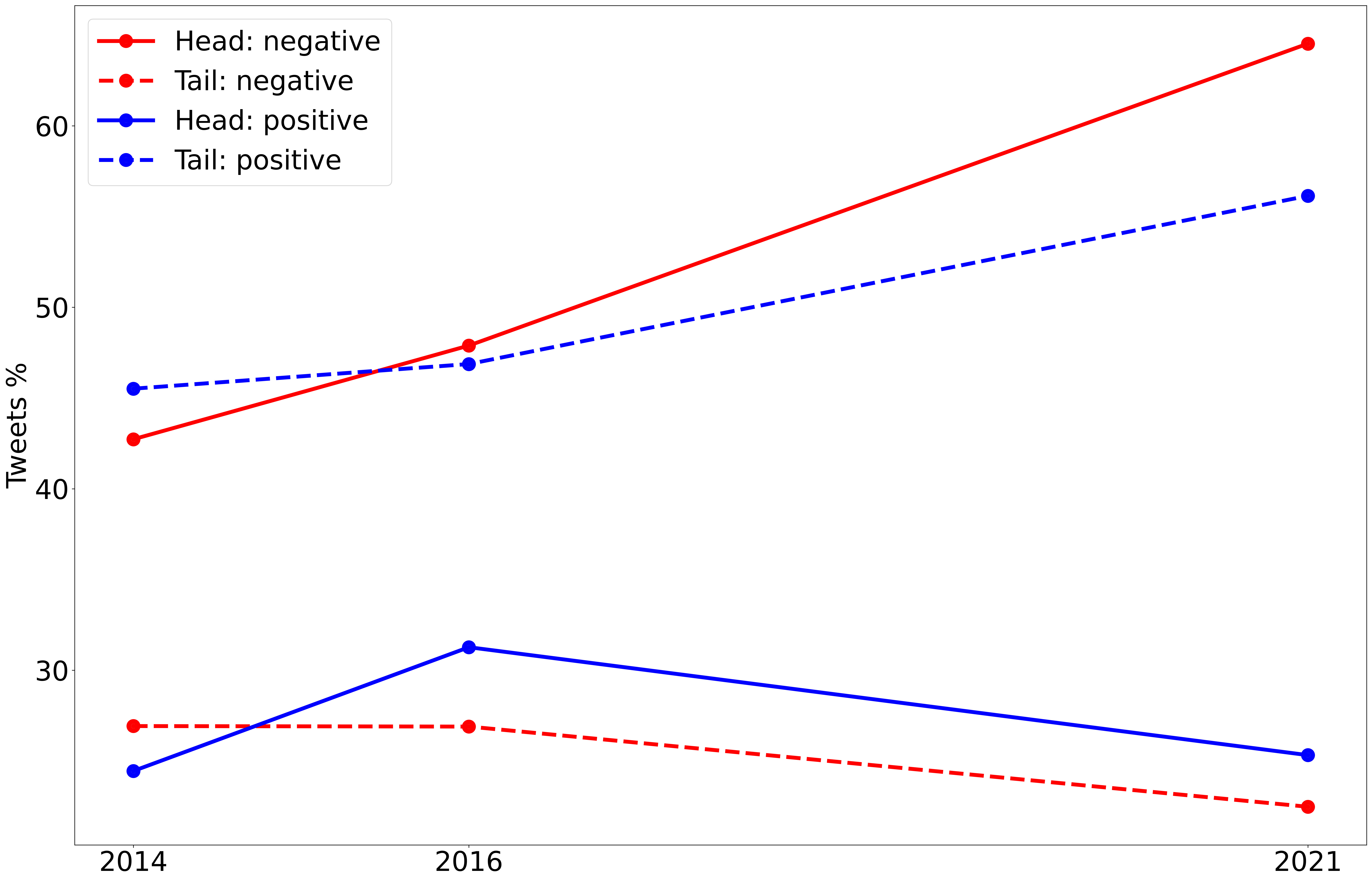}
\caption{Sentiment in `Head' \& `Tail' for tweets from the UK 2014, 2016 and 2021 parliaments.}
\label{fig:temporal_comparison}
\end{figure}

Continuing our analysis, we explore whether the tendency where politicians' negatively charged tweets are more influential is constant through time. To this end, our UK (\textit{2021 Main Dataset}) (Section \ref{sec:data_2021}) along with the \textit{UK 2014 \& 2016} dataset (Section \ref{sec:temporal_data}) are utilised\footnote{Only the UK parliament is considered due to the lack of resources for the past Greek \& Spanish parliaments.}. Figure \ref{fig:temporal_comparison} displays the fluctuation of sentiment (Negative and Positive) in tweets from MPs of the UK's parliament through time.

Again, the tweets are separated in `Head' and `Tail' based on their `popularity'. When considering the `Head' of the distribution, it is clear that tweets with negative sentiment polarity are more numerous than those with positive sentiment throughout the three years studied. As a possibly worrying trend, we can observe how the negativity of tweets in the `Head' grows over time, with a 65\% of all 2021 MP tweets being negative. 
On the other hand, in the distribution of tweets for the `Tail'  the opposite stands, where positively charged tweets outnumber negative tweets by a large margin in all three years, further confirming the main trends discussed in Section \ref{sec:sentvir}.

\paragraph{Discussion} Even if the trend is clearly negative, the idiosyncrasy of each year could potentially explain this trend. For instance, the large discrepancy between Positive and Negative sentiment both in the `Head' and `Tail' of the distribution that is observed on 2021 could be justified due to the Coronavirus pandemic which affected the UK during that time. It is also interesting to note the general increase in negatively charged tweets from 2014 to 2016, 5\% which could be justified due to the talks that took place that year in relation to Brexit and the eventual referendum that took place on June 2016. Further investigation should be required to explain these sociological aspects, not studied in our quantitative research.

\subsection{Sentiment Analysis Models Consistency}\label{sec:sentiment_analysis_models}

In order to ensure the validity of our results, a comparison is made between our selected model (multilingual `XLM-T-Sent') 
and the best performing model for English, `Bertweet-Sent' (see Table \ref{tab:f1_averages} for our main sentiment analysis results). Using the 2021 UK dataset as a test ground, the two models agree on the classification of 80\% of all tweets while reaching a 0.68 agreement score (Cohen's kappa). Besides their overall similar performance, it is important to note that the general trend where `popular' tweets tend to have a bigger network penetration still stands when using `Bertweet-Sent'. More specifically, when considering the `Head' of the dataset tested, both models indicate that the majority of the tweets convey negative sentiment, with 65\% and 62\% of tweets, for `multilingual XLM-T-Sent' and `Bertweet-Sent' respectively, being classified as negative. Similarly, when inspecting the `Tail' of the data, again the models seem to be in agreement with the `multilingual XLM-T-Sent' classifying  22\% of tweets as Negative  and 56\% as positive, while `Bertweet-Sent' classifies 18\% as Negative and 52\% as Positive. The above results, provide further evidence to the robustness of the sentiment analysis classifier.

\section{General Discussion \& Summary of Findings}
\label{sec:summary}

In this section we summarise the main findings of the paper with respect to the research questions (Section \ref{sec:RQsummary}), contextualise them in the light of the control experiments (Section \ref{sec:controlsummary}), and add a general discussion of the implications of the research (Section \ref{sec:discussionsummary}). 

\subsection{Research Questions}
\label{sec:RQsummary}
Having tested several machine learning models and conducted multiple sentiment analysis experiments, we attempt to answer the research questions we set in Section \ref{sec:rqs}.

\paragraph{Sentiment analysis classifier (RQ1)} When considering which is the best sentiment analysis classifier for our particular use case, language models such as `Bertweet-Sent' and `XLM-T-Sent' prove to be the most suitable. As shown in our evaluation (Section \ref{sec:models_evaluation}), language models outperform lexicon-based approaches and traditional machine learning models. However, there is no clear winner when selecting between language models. In our use case, we decided to use `XLM-T-Sent' for most of our experiments as its performance is consistent for all languages studied while also achieving the best scores for Greek and Spanish. Finally, as our dataset contains tweets in other languages too, i.e., Turkish, Catalan, Welsh, using a multilingual model like `XLM-T-Sent' is more convenient to ensure that minority languages are also represented. On the other hand, if our dataset was made only by English tweets then our results indicate that `Bertweet-Sent' would have been a better choice.

\paragraph{Virality and sentiment correlation (RQ2)} Similarly to what previous research has indicated, our experiments confirm that the sentiment of tweets from MPs is an important factor for predicting their popularity. By utilising the sentiment output of `XLM-T-Sent' we performed statistical tests (Chi-square and Kruskal-Wallis H-test) which show that there is a statistically significant difference in the distribution of retweets between positive and negative sentiment (being negative tweets more directly correlated with popularity). These results are supported by the Negative Binomial and Poisson regression models that were tested while using as a dependent variable the retweet count and the sentiment as a predictor. The regression results also reveal that negative sentiment has a larger impact on the tweet's network penetration ability. More details about the regression analysis can be found in Section \ref{sec:sentvir}. 

\paragraph{Sentiment differences in parliaments (RQ3)} We continued our analysis with the aim to explore differences in sentiment of tweets made by different parties. Our results reveal that tweets made by the governing parties tend to be more positive in contrast to tweets made by the opposition parties which are in general more negatively charged (Section \ref{sec:opposition}). This behaviour stands across all of the three main parliaments studied (UK, Spain and Greece).

Moreover, after investigating potential differences in sentiment between the main and devolved parliaments (Section \ref{sec:devolved}) we conclude that even though the overall distribution may differ (i.e., Basque parliament tweets being mostly negative) there is a consistent trend across all parliaments where negatively charged tweets are more popular than tweets conveying positive sentiment.

\subsection{Control experiments} 
\label{sec:controlsummary}
To test the robustness of our results, several control experiments were performed. Firstly, we considered different popularity metrics (Section \ref{sec:popularity_metrics}) including the retweets to followers ratio and the retweets to average user's retweets. Our results indicate that the same general trends, where negatively charged tweets spread further, are consistent independently of which popularity metric is taken into account. 

Potential temporal changes were also explored (Section \ref{sec:temporal_analysis}) by comparing tweets popularity and sentiment from the 2014 and 2016 UK's parliaments. Our experiment confirms that tweets displaying negative sentiment have a bigger network penetration in every year studied while also revealing a trend where the percentage of negatively charged tweets increases through time.

A comparison between  politicians and public twitter accounts was also performed (Section \ref{sec:mps_vs_general}). Our analysis indicate that tweets from public accounts tend to be more positive. However, when considering only tweets from verified accounts differences in the retweet behaviour between countries were identified where in Spain negatively charged tweets were more popular in contrast to UK and Greece where tweets conveying positive sentiment were more retweeted.

Finally, we compare the performance of our model (`XLM-T-Sent') with the best performing English model on the 2021 UK dataset (`Bertweet-Sent') (Section \ref{sec:sentiment_analysis_models}). Our results show that the models reach a high percentage of agreement (80\% agreement on all tweets) and more importantly `Bertweet-Sent' results confirm that negatively charged tweets are in general more popular than tweets conveying positive sentiment.

\subsection{General Discussion}
\label{sec:discussionsummary}

Even though in recent years pre-trained language models have been successfully used on various tasks, when dealing with political communication in Twitter, relevant literature suggests a lack of their utilisation. For our experiments we do take advantage of these state-of-the-art language models along with the data manually annotated to improve our sentiment analysis results. This enables us to reliably answer complex research questions at a large scale.  Moreover, we establish a clear and robust experimental setting that is easy to replicate and apply on different countries and languages. Our analysis confirms the initial hypothesis that politicians tweets charged with negative sentiment are more popular  independently of the country and language studied.

Our results also shed light on the trend where politicians messages tend to be more negative (Section \ref{sec:temporal_analysis}) in recent times, while also highlighting the polarised political environments present where compromises and agreement between parties appear to be rare (Section \ref{sec:opposition}).

It is beyond the scope of this study to consider the wider implications of this trend towards increased negativity in online political communication, or what interventions might mitigate it towards more positive discourse. However, the tools and approach that we have taken can provide a `dashboard' to monitor and detect any changes in this trend across countries, political spheres, and languages over time, including changes in response to any attempts to mitigate the negative discourse.

We offer both our methodology and released multilingual models in the hope that these can be leveraged in future work for subsequent large-scale sociopolitical studies, including other topics such as studying public reaction to climate change and natural disasters \cite{ContrerasMojica2022}. Finally, our methodology can be easily extended to languages not studied in this work by either applying the sentiment model provided directly or by gathering a small annotated corpus and further finetuning the classifier for more precise results.

\section{Conclusion}
We have presented an analysis of the relation between sentiment and virality when considering politicians' tweets. By performing an exhaustive search for a successful sentiment classifier we obtained a robust multilingual model capable of accurately identifying sentiment in politicians' tweets. This is achieved by utilising state-of-the-art transformer-based language models, which we also finetune to the domain-specific task at hand. Both the model used in our analysis and the collected dataset of manually annotated tweets used for training and evaluation are made publicly available.\footnote{https://github.com/cardiffnlp/politics-and-virality-twitter}

Our analysis indicates that there is a strong relationship between the sentiment and the popularity of politicians' tweets, with negatively charged tweets displaying a larger network penetration than tweets conveying positive sentiment. This phenomenon seems to be consistent across all three sovereign countries analysed (i.e., Greece, Spain and United Kingdom). 
Our findings are further verified by the control experiments performed. 
Among these control experiments, a temporal analysis suggests that the trend of negative tweets being more influential is consistent and getting more pronounced over time. Finally, while tangential to the main research questions, we observe a clear distinction between government and opposition parties irrespective of their ideology, with government parties and leaders being more positive overall.

Future work could follow our methodology to extend our analysis with additional parliaments and a wider set of time periods taken into consideration. Furthermore, a more fine-grained classification process, where sentiment is classified on a scale, e.g., 1--5 or considering various aspects, might be useful to discover subtler relations between sentiment and virality that are unseen when using a three class classification (negative, positive, neutral). Finally, we hope that both our methodology and released multilingual models can be leveraged in future work for subsequent large-scale sociological studies, including other topics.

\paragraph{Ethics Statement}
In this paper we explore the relation between sentiment and virality in politicians' tweets with the aim to acquire a better understanding of how politicians utilize twitter and their relations with the public. As the aim of this work was to identify general trends, only aggregate statistics are displayed and no attempt is made to identify or focus upon individual MPs (with the exception of the public party leaders). Our experiments therefore respect the privacy of individuals and also comply with Twitter's policies (https://developer.twitter.com/en/developer-terms). At the same time, as the main focus of our analysis deals with public figures (MPs) the content of our analysis is by definition addressed to the general population. All of the data used in the experiments are public and accessible through Twitter and are also made available in our repository where we share the tweet IDs that are used. 

We note that some of our experiments identify specific groups and aim to differentiate between them. Mainly, a comparison is made between political parties (Figure \ref{fig:2021_parties}) and their usage of sentiment in tweets; however, our analysis remains neutral, considering only the objective differences between governing and opposition parties, irrespective of their political stances. Nonetheless, we are aware that our analysis, and potential use of our dataset by others, can be potentially used in a politically-biased way.

\section*{Acknowledgements}
Jose Camacho-Collados is supported by a UKRI Future Leaders Fellowship.

We also acknowledge the help of Ángela Collados Aís, Carla Perez Almendros, Dimitra Mavridou, Mairi Antypa, David Owen, Matthew Redman and David Humphreys in the annotation task.

 \bibliographystyle{elsarticle-num} 
\bibliography{bibliography}

\appendix

\section{Data Collection}
Table \ref{tab:tweets_collected} displays the number of tweets collected from each parliament for the main time period studied, January to December 2021.

Table \ref{tab:control_stats_devolved} displays text statistics extracted from tweets of the devolved parliaments of N.Ireland, Scotland, Wales, Catalonia and Basque country. The percentage of tweets that include emojis, hashtags, mentions, and URLs as well as the percentage of upper case characters are displayed. In general, the devolved parliaments seem to follow a similar trend to their respective main bodies (Section \ref{sec:data_2021}, Table \ref{tab:control_stats}).

\begin{table*}[ht]
\begin{adjustbox}{width=\textwidth,center}
\begin{tabular}{|l|l|l|l|l|l|l|l|l|l|l|l|l|l|}
\hline
             & Jan & Feb & Mar & Apr & May   & Jun  & Jul  & Aug & Sep & Oct & Nov & Dec & \textbf{Total} \\ \hline
UK          & 47,296 & 47,606 & 38,965 & 29,336 & 34,215 & 31,081 & 31,970 & 22,044 & 27,509 & 29,557 & 31,330 & 29,026 & 399,935 \\ \hline
-Scotland  & 11,282 & 13,175 & 8,988  & 6,622  & 7,890  & 5,600  & 3,856  & 5,295  & 6,285  & 5,301  & 6,427  & 6,241  & 86,962  \\ \hline
-Wales     & 5,381  & 5,462  & 3,903  & 3,386  & 3,874  & 3,013  & 3,191  & 2,393  & 2,864  & 2,826  & 3,410  & 3,084  & 42,787  \\ \hline
-N.Ireland & 5,945  & 6,788  & 5,640  & 4,579  & 5,127  & 6,138  & 4,798  & 3,484  & 4,947  & 4,505  & 4,452  & 5,391  & 61,794  \\ \hline
Spain        & 20,119   & 25,343    & 18,407 & 18,073 & 16,938 & 16,167 & 14,732 & 11,595  & 15,376     & 14,753   & 13,823    & 13,175    & 198,501         \\ \hline
-Catalonia & 9,512  & 8,386  & 6,463  & 5,537  & 6,389  & 6,030  & 6,047  & 3,729  & 5,940  & 5,856  & 5,376  & 5,255  & 74,520  \\ \hline
-Basque    & 2,103  & 2,299  & 1,478  & 1,287  & 1,230  & 989   & 685   & 843   & 1,040  & 1,118  & 1,161  & 1,091  & 15,324  \\ \hline
Greece      & 7,234  & 9,298  & 7,421  & 7,136  & 6,345  & 6,140  & 6,239  & 4,300  & 6,060  & 5,458  & 5,941  & 5,083  & 76,655  \\ \hline
\textbf{ALL} & 108,872  & 118,357   & 91,265 & 75,956 & 82,008 & 75,158 & 71,518 & 53,683  & 70,021     & 69,374   & 71,920    & 68,346    &    956,478            \\ \hline
\end{tabular}
\end{adjustbox}
\caption{Number of MP tweets collected through 2021.}
\label{tab:tweets_collected}
\end{table*}

\begin{table}[ht]
    \centering
    \begin{tabular}{lccccc}
    \textbf{Country} & \textbf{\#} & \textbf{@} & \textbf{emoji} & \textbf{url} & \textbf{up/low} \\ \hline
    N.Ireland               & 0.04        & 0.39       & 0.25           & 0.45         & 0.11            \\ \hline
    Scotland            & 0.04        & 0.30       & 0.29           & 0.54         & 0.11            \\ \hline
    Wales           & 0.04        & 0.26       & 0.31           & 0.59         & 0.12            \\ \hline
    Catalonia           & 0.07        & 0.40       & 0.33           & 0.69         & 0.11            \\ \hline
    Basque           & 0.05        & 0.33      & 0.27           & 0.63         & 0.09            \\ \hline
    \textbf{Overall} & 0.05        & 0.34       & 0.29           & 0.58         & 0.11           
    \end{tabular}
    \caption{Percentage of tweets that include  hashtags (\#), mentions (@), emojis, and URLs in \textit{2021 Devolved Dataset}. The upper to low case ratio (upp/low) average of tweets texts is also reported.}
    \label{tab:control_stats_devolved}
\end{table}

\section{Extended classification results}
Table \ref{tab:f1_averages} displays the macro average F1 scores along with the average F1 scores of positive and negative classes (F1\textsuperscript{PN}) for all the experiments performed. The experiments were performed by using both a cross-validation setting (CV) and a train/test split approach. For the train/test approach the set of tweets that was annotated by all coders was used (approximately 100 tweets for each language) while the rest of the data was used for training (approximately 900 tweet for each language) (Section \ref{sec:annotation}). The results indicate the \textit{XLM-T-Sent-multi} performs consistently in both the CV and train/test settings and achieves satisfactory performance particularly for Spanish. 

Finally, our results demonstrate the importance that a specialised corpus can have when training the models. In particular, when looking at the $F1$ scores for the `CV' setting the models display an increased performance when trained further with the `In' domain, labelled data collected (Section \ref{sec:annotation}). The performance of `XLM-T-Sent', `TweetEval-Sent' and `Bertweet-Sent' models is increased (or at worst stays the same), across all languages tested when the `In' domain data are added to their training corpus. 

\begin{table*}[hbp!]
\begin{adjustbox}{width=\textwidth,center}
\begin{tabular}{|l|l|l|c|c|c|c|c|c|c|c|c|c|c|c|c|c|}
\hline
\multirow{3}{*}{Type} & \multirow{3}{*}{Model} & \multirow{3}{*}{\begin{tabular}[c]{@{}l@{}}Train \\ Lang\end{tabular}} & \multicolumn{2}{c|}{\multirow{2}{*}{Train domain}} & \multicolumn{4}{c|}{United Kingdom} & \multicolumn{4}{c|}{Spain} & \multicolumn{4}{c|}{Greece} \\ \cline{6-17} 
 &  &  & \multicolumn{2}{c|}{} & \multicolumn{2}{c|}{F1} & \multicolumn{2}{c|}{F1\textsuperscript{PN}} & \multicolumn{2}{c|}{F1} & \multicolumn{2}{c|}{F1\textsuperscript{PN}} & \multicolumn{2}{c|}{F1} & \multicolumn{2}{c|}{F1\textsuperscript{PN}} \\ \cline{4-17} 
 &  &  & Out & In & CV & split & CV & split & CV & split & CV & split & CV & split & CV & split \\ \hline
\multirow{10}{*}{\rot{Multingual}} & \multirow{3}{*}{XLM-T} & Mono &  & \checkmark & 70 & 72 & 78 & 75 & 67 & 66 & 80 & 80 & 74 & 76 & 73 & \textbf{74} \\ \cline{3-17} 
 &  & Mono & \checkmark & \checkmark & 74 & 65 & 80 & 76 & 70 & 76 & 80 & 81 & 75 & 66 & 73 & 61 \\ \cline{3-17} 
 &  & Multi &  & \checkmark & 75 & 67 & 81 & 77 & \textbf{72} & 78 & 80 & 84 & 76 & \textbf{77} & 75 & 75 \\ \cline{2-17} 
 & \multirow{3}{*}{XLM-T-Sent} & Mono & \checkmark &  & 72 & 72 & 78 & 79 & 68 & 71 & 77 & 82 & 65 & 69 & 64 & 68 \\ \cline{3-17} 
 &  & Mono & \checkmark & \checkmark & 74 & 67 & 80 & 74 & 70 & 70 & \textbf{81} & 78 & 77 & 76 & 75 & \textbf{74} \\ \cline{3-17} 
 &  & Multi & \checkmark & \checkmark & 74 & 68 & 80 & 75 & 70 & \textbf{79} & \textbf{81} & \textbf{87} & 76 & 70 & 77 & 70 \\ \cline{2-17} 
 & \multirow{4}{*}{XLM-R} & Mono &  & \checkmark & 69 & 72 & 77 & 75 & 60 & 65 & 75 & 80 & 74 & 75 & 73 & 73 \\ \cline{3-17} 
 &  & Mono & \checkmark & \checkmark & 70 & 67 & 74 & 74 & 66 & 65 & 80 & 78 & 74 & 74 & 73 & 70 \\ \cline{3-17} 
 &  & Multi &  & \checkmark & 74 & 66 & 80 & 73 & 69 & 70 & 80 & 80 & \textbf{78} & 72 & \textbf{78} & 70 \\ \cline{3-17} 
 &  & Multi & \checkmark & \checkmark & 70 & 57 & 78 & 67 & 67 & 70 & 76 & 80 & 76 & 72 & 76 & 72 \\ \hline
\multirow{6}{*}{\rot{Monolingual}} & \multirow{2}{*}{RoBERTa-Base} & Mono &  & \checkmark & 68 & 73 & 78 & 77 & 64 & 70 & 78 & 86 & 56 & 57 & 50 & 52 \\ \cline{3-17} 
 &  & Mono & \checkmark & \checkmark & 75 & 70 & 80 & 75 & 70 & 78 & 80 & \textbf{87} & 56 & 62 & 49 & 58 \\ \cline{2-17} 
 & \multirow{2}{*}{TweetEval-Sent} & Mono & \checkmark &  & 73 & 72 & 80 & 79 & \multicolumn{8}{c|}{\multirow{4}{*}{}} \\ \cline{3-9}
 &  & Mono & \checkmark & \checkmark & 76 & 67 & 81 & 74 & \multicolumn{8}{c|}{} \\ \cline{2-9}
 & \multirow{2}{*}{Bertweet-Sent} & Mono & \checkmark &  & 77 & \textbf{76} & \textbf{83} & \textbf{83} & \multicolumn{8}{c|}{} \\ \cline{3-9}
 &  & Mono & \checkmark & \checkmark & \textbf{78} & 75 & 82 & 78 & \multicolumn{8}{c|}{} \\ \hline
- & \multirow{2}{*}{SVM} & Mono &  & \checkmark &  33 &  36 & 50 & 47 & 40 & 40  & 60 & 61 & 60 & 63 & 56 & 60 \\ \cline{1-1}  \cline{3-17} 
- &  & Mono & \checkmark & \checkmark &  47 & 47 & 61 & 57 & 42 & 38 & 58 & 57 & 52 & 55 &  44 &  51 \\ \hline
- &  \multirow{2}{*}{LSTM} & Mono  &  & \checkmark & 52  & 48 &  66 & 55 & 51 & 48 & 60 & 61 & 59 & 62 & 56 &  61\\ \cline{1-1} \cline{3-17} 
- & & Mono & \checkmark & \checkmark & 58 & 55 & 67 & 63 & 49  & 42  & 60 & 56 &  59 &60  & 57 & 59 \\ \hline
- & VADER &  &  &  & 55 & 47 & 64  & 61  & 54 & 57 & 63 & 66 & 57  & 51 & 58 & 53  \\ \hline
- & Baseline &  &  &  & 22 & 22 & 33 & 33 & 20 & 21 & 30 & 32 & 17 & 20 & 27 & 30 \\ \hline
\end{tabular}
\end{adjustbox}
\caption{Average F1 scores along with the average of F1 scores of positive and negative classes (F1\textsuperscript{PN}) when trained/evaluated with cross-validation (CV) and on the $\sim$900/100 train/test split. The results displayed are the averages of three runs. Training domain indicates the data used for training the models. `In' domain indicates models trained on the labelled data collected (Section \ref{sec:annotation}), `Out' indicates the model has been finetuned on other sentiment analysis datasets (see Section \ref{sec:expsetting}). Finally, `Mono' indicates that the model was trained with data from only one language, while `Multi' shows models that were trained with data from all available languages.}
\label{tab:f1_averages}
\end{table*}

\section{Extended regression results}
Table \ref{tab:poisson_results} displays the results (coefficients) of the Poisson and zero inflated Poisson regression models while using the retweet count as a dependent variable. Similarly to the \textit{NBR} models used (Section \ref{sec:sentvir}: \textit{Regression Analysis}) both Poisson models indicate that the existence of negative sentiment influences the retweet count greatly than that of positive sentiment.

\begin{table*}[hbp!]
\centering
\begin{tabular}{|l|c|c|c|c|}
\hline
\textbf{variable} & \textbf{Overall} & \textbf{UK} & \textbf{Spain} & \textbf{Greece} \\ \hline
const             & 72.8914*         & 75.9854*    & 96.6305*       & 96.6305*        \\ \hline
emojis            & -5.1085*         & -7.9482*    & -9.7004*       & -9.7004*        \\ \hline
has\_url          & -1.3613          & -17.524*    & 26.56*         & 26.56*          \\ \hline
hashtags          & -20.0335*        & -7.811*     & -41.0476*      & -41.0476*       \\ \hline
mentions          & -41.0511*        & -40.4977*   & -63.0461*      & -63.0461*       \\ \hline
max\_score        & -43.16*          & -42.5478*   & -39.615*       & -39.615*        \\ \hline
\end{tabular}
\caption{Coefficients for the OLS model when using tweets from the combined 2021 UK, Spain, Greece parliaments and for each individual parliament. Dependent variable is the retweet count while XLM-T's softmax scores (negative and positive sentiment) are utilised (sent score) as independent variables.  $*$ indicates p-value $< 0.05$ }
\label{tab:ols_results}
\end{table*}

Table \ref{tab:ols_results} shows the coefficients for each predictor when using the softmax scores of the positive and negative sentiments of the \textit{XLM-T} model instead of the final label. For each entry the highest absolute softmax score is considered and is assigned positive or negative sign based on the sentiment. The results indicate, i.e., negative coefficient, that the "more" negative a tweet the more is retweeted.

\begin{table*}[hbp!]
\begin{adjustbox}{width=\textwidth,center}
\begin{tabular}{|l|cc|cc|cc|cc|}
\hline
\multirow{2}{*}{\textbf{variable}} &
  \multicolumn{2}{c|}{\textbf{Overall}} &
  \multicolumn{2}{c|}{\textbf{UK}} &
  \multicolumn{2}{c|}{\textbf{Spain}} &
  \multicolumn{2}{c|}{\textbf{Greece}} \\ \cline{2-9} 
 &
  \multicolumn{1}{c|}{\textbf{Poisson}} &
  \textbf{zero Poisson} &
  \multicolumn{1}{c|}{\textbf{Poisson}} &
  \textbf{zero Poisson} &
  \multicolumn{1}{c|}{\textbf{Poisson}} &
  \textbf{zero Poisson} &
  \multicolumn{1}{c|}{\textbf{Poisson}} &
  \textbf{zero Poisson} \\ \hline
const &
  \multicolumn{1}{c|}{3.4035*} &
  4.0809* &
  \multicolumn{1}{c|}{3.7006*} &
  4.4486* &
  \multicolumn{1}{c|}{3.61*} &
  4.1586* &
  \multicolumn{1}{c|}{2.3372*} &
  2.8278* \\ \hline
emojis &
  \multicolumn{1}{c|}{-0.193*} &
  -0.1769* &
  \multicolumn{1}{c|}{-0.3277*} &
  -0.2978* &
  \multicolumn{1}{c|}{-0.2114*} &
  -0.1946* &
  \multicolumn{1}{c|}{-0.3298*} &
  -0.3639* \\ \hline
has\_url &
  \multicolumn{1}{c|}{0.013*} &
  -0.3695* &
  \multicolumn{1}{c|}{-0.3798*} &
  -0.8422* &
  \multicolumn{1}{c|}{0.41*} &
  0.0827* &
  \multicolumn{1}{c|}{-0.1871*} &
  -0.305* \\ \hline
hashtags &
  \multicolumn{1}{c|}{-0.5219*} &
  -0.6281* &
  \multicolumn{1}{c|}{-0.4001*} &
  -0.5168* &
  \multicolumn{1}{c|}{-0.5878*} &
  -0.6266* &
  \multicolumn{1}{c|}{-0.3372*} &
  -0.41* \\ \hline
mentions &
  \multicolumn{1}{c|}{-1.0543*} &
  -0.7944* &
  \multicolumn{1}{c|}{-1.4615*} &
  -0.8382* &
  \multicolumn{1}{c|}{-0.929*} &
  -0.8291* &
  \multicolumn{1}{c|}{-0.9082*} &
  -0.7696* \\ \hline
neg &
  \multicolumn{1}{c|}{1.4129*} &
  1.1709* &
  \multicolumn{1}{c|}{1.3014*} &
  1.0566* &
  \multicolumn{1}{c|}{1.1818*} &
  1.0028* &
  \multicolumn{1}{c|}{0.9232*} &
  0.6859* \\ \hline
pos &
  \multicolumn{1}{c|}{0.2295*} &
  0.1629* &
  \multicolumn{1}{c|}{-0.1486*} &
  -0.1855* &
  \multicolumn{1}{c|}{0.5513*} &
  0.4503* &
  \multicolumn{1}{c|}{0.3229*} &
  0.2315* \\ \hline
\end{tabular}
\end{adjustbox}
\caption{Coefficients for the Poisson Regression and zero inflated Poisson Regression when using tweets from the combined 2021 UK, Spain, Greece parliaments and for each individual parliament. Dependent variable is the retweet count while the independent variables \textit{neg} and \textit{pos} indicate the sentiment according to XLM-T results. $*$ indicates p-value $< 0.05$ }.
\label{tab:poisson_results}
\end{table*}

\end{document}